\documentclass[10pt,twocolumn,letterpaper]{article}

\usepackage[pagenumbers]{cvpr} 
\usepackage{multirow}

\newcommand{\myparagraph}[1]{\smallskip\vspace{-0.02in}\noindent\textbf{#1}}

\definecolor{cvprblue}{rgb}{0.21,0.49,0.74}
\usepackage[pagebackref,breaklinks,colorlinks,allcolors=cvprblue]{hyperref}
\usepackage[accsupp]{axessibility}

\def\paperID{7478} 
\def\confName{CVPR}
\def\confYear{2025}

\title{MERGE: Multi-faceted Hierarchical Graph-based GNN for Gene Expression Prediction from Whole Slide Histopathology Images}

\author{
Aniruddha Ganguly\textsuperscript{1}\thanks{Email: aniganguly@cs.stonybrook.edu}\hskip 1em
Debolina Chatterjee\textsuperscript{2}\hskip 1em
Wentao Huang\textsuperscript{1}\hskip 1em
Jie Zhang\textsuperscript{2}\hskip 1em
Alisa Yurovsky\textsuperscript{1}\hskip 1em \\
Travis Steele Johnson\textsuperscript{2}\hskip 1em
Chao Chen\textsuperscript{1}\hskip 1em
\\
\textsuperscript{1}Stony Brook University, NY, USA \hskip 1em
\\
\textsuperscript{2}Indiana University School of Medicine, IN, USA \hskip 1em
}

\begin{document}
\maketitle
\begin{abstract}
Recent advances in Spatial Transcriptomics (ST) pair histology images with spatially resolved gene expression profiles, enabling predictions of gene expression across different tissue locations based on image patches. This opens up new possibilities for enhancing whole slide image (WSI) prediction tasks with localized gene expression. However, existing methods fail to fully leverage the interactions between different tissue locations, which are crucial for accurate joint prediction. To address this, we introduce \textbf{MERGE} (Multi-faceted hiErarchical gRaph for Gene Expressions), which combines a multi-faceted hierarchical graph construction strategy with graph neural networks (GNN) to improve gene expression predictions from WSIs. By clustering tissue image patches based on both spatial and morphological features, and incorporating intra- and inter-cluster edges, our approach fosters interactions between distant tissue locations during GNN learning. 
As an additional contribution, we evaluate different data smoothing techniques that are necessary to mitigate artifacts in ST data, often caused by technical imperfections. We advocate for adopting gene-aware smoothing methods that are more biologically justified. 
Experimental results on gene expression prediction show that our GNN method outperforms state-of-the-art techniques across multiple metrics. 

\vspace{-0.25in}
\end{abstract}
\section{Introduction}
\label{sec:intro}

\begin{figure}
    \centering
    \includegraphics[width=1\linewidth]{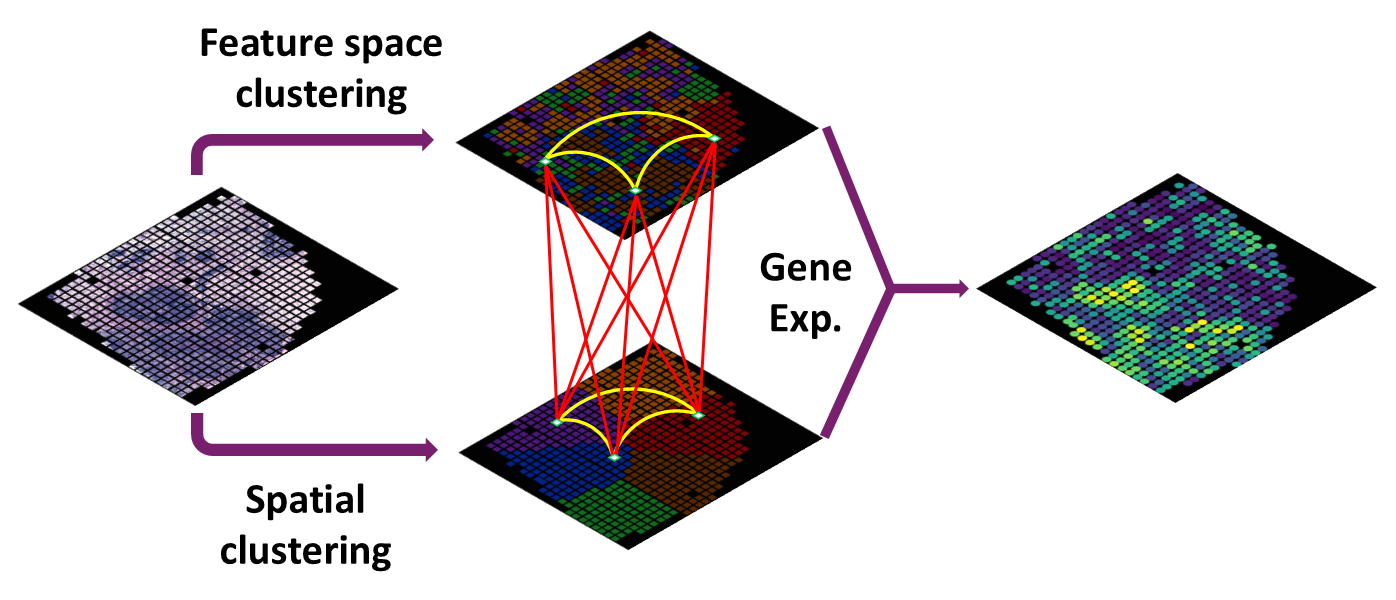}
    \caption{Combining image feature-space clustering with spatial clustering to build a multi-faceted hierarchical graph leads to spatial gene expression prediction. \textbf{Left:} Patches extracted from a WSI. \textbf{Middle:} Edges within and across clusters allows us to model short and long range interactions among patches. \textbf{Right:} This allows our GNN model to make robust predictions.}
    \label{fig:teaser}
    \vspace{-0.25in}
\end{figure}

Spatial Transcriptomics (ST) entails the quantification of mRNA expression across a set of genes in a tissue sample, splitting the sample into `spots'. Previous techniques like bulk RNA sequencing quantify gene expressions for an entire tissue sample, thereby sacrificing both the heterogeneity of the sample and the spatial context. Single-cell RNA sequencing (scRNA-seq), however, captures the heterogeneity at the cell-level without preserving the spatial context. Recent ST methods including Visium \cite{doi:10.1126/science.aaf2403}, MERFISH \cite{doi:10.1126/science.aaa6090}, seqFISH+ \cite{eng2019transcriptome}, STARmap \cite{doi:10.1126/science.aat5691}, smFISH \cite{Codeluppi276097}, and Targeted ExSeq \cite{doi:10.1126/science.aax2656} have bridged these gaps. In these methods, a spot can refer to a tissue segment containing multiple cells \cite{doi:10.1126/science.aaf2403}, a single cell \cite{eng2019transcriptome}, or a sub-cellular region \cite{vickovic2019high}. Despite the tremendous scientific premises, ST technology is very expensive due to the necessity of domain expertise, specialized equipment, and expensive reagents.

To circumvent the challenge in acquiring ST data, one can learn to predict gene expression profiles from histology images. This provides the opportunity to estimate localized gene expression on a whole slide image (WSI), which in turn can potentially benefit many image-based downstream tasks.
Most recent works on this direction \cite{he2020integrating, Pang2021.11.28.470212, xie2023spatially, yang2023exemplar, 10.1093/bib/bbac297, chung2024accurate} train on the Visium ST data~\cite{doi:10.1126/science.aaf2403}, in which a tissue sample is divided into thousands of regularly distributed \emph{spots}. Each spot has both the histology image patch and the corresponding gene expression profile. Therefore, we will use `spot' to refer to image patches and their corresponding gene expressions interchangeably throughout this paper. Past methods train on these image-gene pairs to learn to predict gene expressions at different locations of a tissue sample based on purely histology images.

This task is challenging. Given already very powerful image representations~\cite{chen2024uni, vorontsov2024foundation, lu2023visual, huang2023visual, lu2024visual}, it is difficult to extract the gene-predicting morphological information that has not yet been used. Another impediment is the presence of artifacts due to the imperfect ST technology. Measurements are often noisy and sparse, with a high dropout rate (i.e., certain genes have zero expression when they should not)~\cite{kharchenko2014bayesian}. To address these challenges, we need to not only explore advanced vision techniques, but also exploit the intrinsic structure of the data; image patches with similar visual patterns and/or spatial proximity should share the relevant information and make joint predictions.

Existing methods have made progress in exploring such patch structure. An earlier work (ST-Net~\cite{he2020integrating}) directly uses CNN features from each patch to predict its gene expression, overlooking the interaction between patches. The image patch representation has been improved by better multi-modal representation learning~\cite{jaume2024hest, xie2023spatially}. Methods~\cite{Pang2021.11.28.470212, yang2023exemplar} have been proposed to explore long-range interactions between patches within a slide, where WSI level contexts have been exploited using advanced learning techniques such as vision transformer (ViT)~\cite{dosovitskiy2020image}. But ViT expects self-attention across patches to be automatically discovered from data, and thus can be negatively affected by scarcity of ST data, heterogeneity of the tissue micro-environment, and the ST data quality limitation outlined above. 
The latest method, TRIPLEX \cite{chung2024accurate}, outperforms previous works by hard coding a much stronger prior into the model. For a single patch, it extracts features from multiple resolutions - the patch itself, its nearby neighbor patches, and the whole slide. However, this prior is very rigid and is not flexible enough to facilitate the discovery and interaction of similar patches separated by long distances. 

To learn long-range interactions while being data-efficient, graph neural networks (GNNs) would be a natural choice. A GNN allows message passing via edges and thus can facilitate long-range interaction between nodes. However, Hist2ST \cite{10.1093/bib/bbac297}, the sole prior GNN approach, uses only one-hop edges, i.e., connecting each patch to its immediate neighbor. Effective long-range interactions are then hard to achieve due to the limited number of layers of a GNN. To build an effective GNN for this task, we need biologically justified long-range edges or \textit{shortcuts} connecting distant patches.

In this paper, we introduce \textbf{MERGE}, a novel GNN method which facilitates targeted long-range interaction, and thus achieves stronger performance in the joint gene expression prediction task. The GNN takes different patches/spots as nodes, using their image features as input, and makes joint predictions of the gene expressions across all patches.
Our key innovation is to leverage prior knowledge to build a multi-faceted hierarchical graph with both short and long-range interactions among patches. 
To accomplish these, we perform both spatial and feature space clustering as seen in \cref{fig:teaser}. Long-range interactions are enabled through \textit{shortcut edges} across clusters. This keeps the graph sparse, but propagates useful information to distant yet biologically relevant nodes in fewer hops. Experimental evidence demonstrates the superiority of our method due to the carefully-crafted multi-faceted hierarchical graph.

Another challenge for learning with ST data is data quality, where the gene expression values are noisy and sparse, and exhibit high dropouts~\cite{doi:10.1126/science.aaf2403, ding2020systematic}. To mitigate this issue, most previous methods employ a single-factor smoothing method that only performs spatial smoothing over an immediate neighborhood. But the sparsity of ST data means spatial smoothing spreads non-zero values to neighboring spots, reducing their magnitude while diffusing their influence. Furthermore, the smoothing is not gene-aware, and retains little correspondence with tissue morphology. To address the issue, we propose to adopt a gene-informed bi-modal smoothing technique \cite{10.1093/bib/bbac116} and demonstrate the superiority of such smoothing method in producing gene expression values that are much better aligned with biology. Our thorough quantitative and visual analysis demonstrate the necessity of adopting such gene-informed smoothing methods for ST data learning, and could be counted as a separate technical contribution.

We summarize our key contributions as follows:
\begin{itemize}
    \item We use a graph neural network to capture local and global dependencies between similar spots, leading to better correlation between predicted gene expressions and the ground truth
    \item We propose a multi-faceted hierarchical graph construction strategy to identify and utilize the dependencies among similar tissue clusters regardless of spatial proximity, leading to more meaningful interactions among nodes.
    \item We demonstrate that using a gene-informed smoothing technique improves gene-tissue correlation substantially and reinforces the long-range similarities and dissimilarities among spots.
\end{itemize}

\section{Related Work}
\label{sec:related}

\begin{figure*}[t]
    \centering
    \includegraphics[width=1\textwidth]{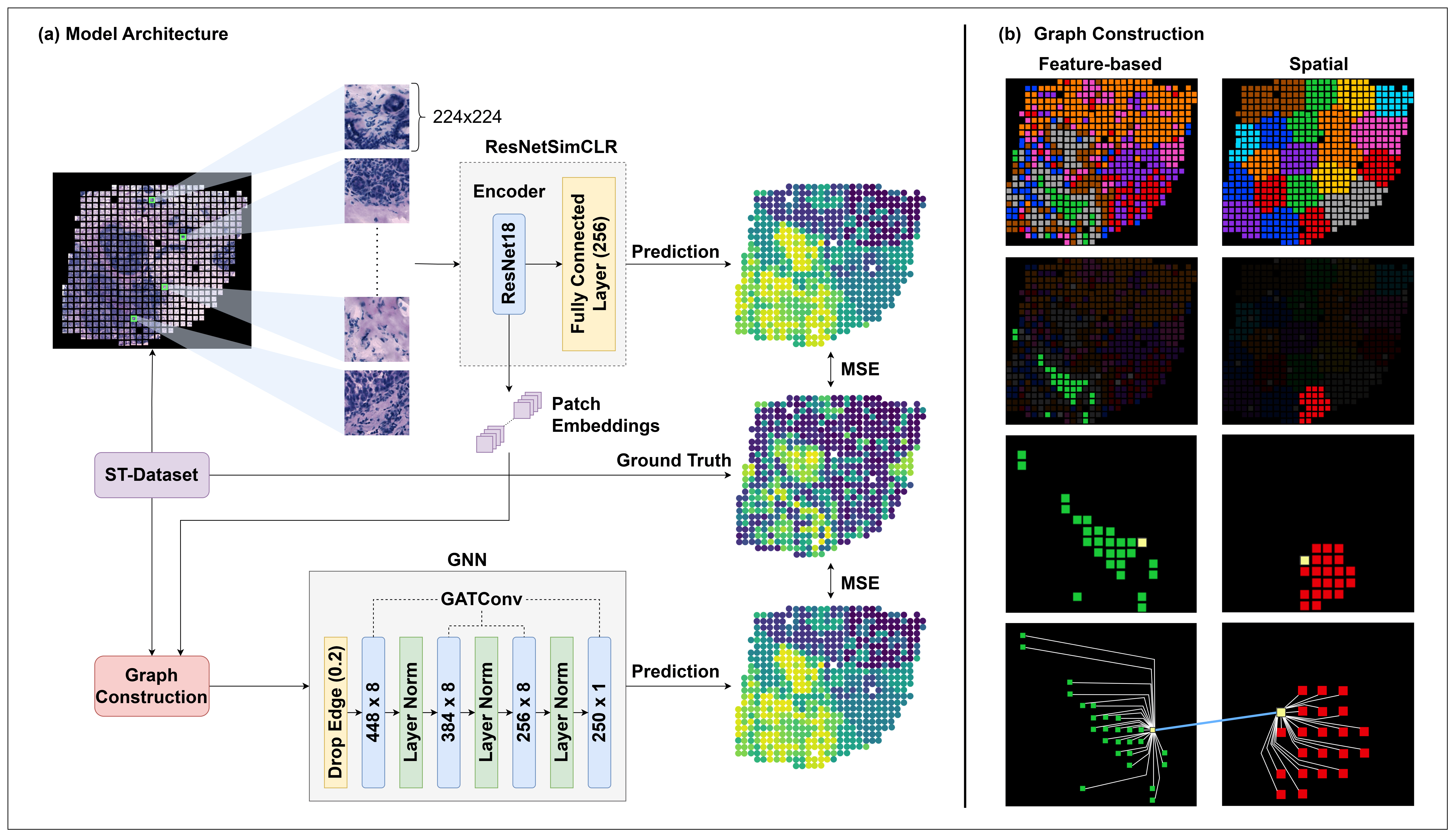}
    \caption{The schematic of MERGE shows the overall workflow of our method. (a) Outlines the architecture of our method. The ResNetSimCLR model is fine-tuned on the gene expression prediction task using MSE loss. The last layer is discarded to yield 256-dimensional feature vectors for the patches. The graph construction step produces the multi-faceted hierarchical graph for our GNN, which is trained on MSE loss. The output of the GNN is a 250-dimensional gene expression vector at each node. (b) Shows the graph construction strategy demonstrated through reduced examples. The left column shows feature space clustering and the right column shows spatial clustering. The internal edges of a cluster are shown in white, while the shortcut edge is shown in blue. The two yellow spots represent the centroid spots of the two clusters.}
    \label{fig:architecture}
\vspace{-0.25in}
\end{figure*}

This section explores several prior studies that are relevant to gene expression prediction using image features. We analyze them based on the context used to extract image features and how effective they are at modeling long-range interactions among similar tissue segments. Additionally, we discuss gene expression smoothing methods that are crucial to our qualitative analysis.

\myparagraph{Gene expression prediction.}
Predicting gene expression from WSI is a well explored topic. But only a few of these studies have worked with spatial gene expression prediction. ST-Net \cite{he2020integrating} was the first study to explore this task, using a DenseNet-121 model \cite{huang2017densely} with a fully connected layer at the end to predict the expression for 250 target genes. This only considers image features of the target spot, but ignores both local and long-range or global contexts. HisToGene \cite{Pang2021.11.28.470212} incorporates positional embeddings alongside patch embeddings to account for positional information in the feature representation for a spot, using Vision Transformers \cite{dosovitskiy2020image} to capture positional dependency among patches. This makes the model aware of positional dependencies on a global scale, but fails to capture feature-space correlations. A slightly different approach is seen in EGN \cite{yang2023exemplar}, where dynamically selected exemplars reinforce gene expression predictions for a target patch. A contrastive learning approach by BLEEP \cite{xie2023spatially} attempts to circumvent the local-vs-global context issue by enhancing image feature extracting using gene expressions. The authors use gene expression embeddings to guide the pre-training phase of a ResNet50 \cite{he2016deep} based patch encoder. This allows better performance than prior methods. Alongside the issue of not being able to effectively capture long-range dependencies, both EGN and BLEEP suffer from the problem of computation-heavy inference. 

The sole use of Graph Neural Networks (GNNs) to predict gene expressions in ST data is seen in Hist2ST \cite{10.1093/bib/bbac297}, where a ConvMixer \cite{trockman2022patches} acts as a feature extractor for the target spot, a transformer captures global spatial dependencies, and a `four nearest neighbors graph' is used in a GraphSAGE network \cite{hamilton2017inductive} to predict gene expressions. While the use of GNNs allows Hist2ST to capture better correlation among patches, there is little to no message passing among distant but similar nodes. 

The state of the art in spatial gene expression prediction is TRIPLEX \cite{chung2024accurate}, which concurrently trains three encoders to extract features from a target patch, its spatial neighborhood (local context) and all patches in the WSI (global context). The extracted features are subsequently fused by another encoder and passed through a linear layer to make gene expression predictions. This approach works much better than all prior methods and captures the global context to a reasonable degree. The fact remains, however, that none of these approaches have been able to effectively model long-range interactions among spots.

\myparagraph{Gene expression smoothing techniques for ST data.}
Transcriptomics data in general exhibits noise and dropout events, with ST suffering from these to a much higher degree~\cite{doi:10.1126/science.aaf2403, ding2020systematic}. As a result, data smoothing for older techniques like scRNA-seq has been extensively studied and are commonplace in transcriptomics based tasks~\cite{hou2020systematic}. Smoothing techniques like SAVER \cite{huang2018saver} and MAGIC~\cite{van2018recovering} are designed for scRNA-seq data and are transcriptome-aware. But they do not take the spatial associations of spots into consideration. Therefore, they do not work as well when applied to ST data. Previous studies on histology-based gene expression prediction have used a spatial smoothing technique where the gene expressions of each spot are replaced with the averages of the expressions of a nine-spot neighborhood including itself. Such spatial smoothing techniques are not transcriptome-aware. More recent techniques like SPCS~\cite{10.1093/bib/bbac116} and EAGS~\cite{10.1093/gigascience/giad097} are simultaneously transcriptome-aware and spatial association-aware. This makes them much more robust compared to single-factor smoothing methods.
\section{Method}
\label{sec:method}

Our method is trained on ST data, but can be applied to any WSI data at inference.
During training, the model takes as input a set of patches and their corresponding gene expressions. We encode each patch into embeddings, and construct a multi-faceted hierarchical graph where each spot is a node. This graph is used in a GNN, which learns to predict gene expressions for each node. The optimization objective is a minimization of the MSE loss over the predicted and ground truth gene expressions. Note that we are jointly predicting on all genes at the same time. In practice, many genes are not expressed, and following previous methods, we select genes with sufficient presence as our prediction targets. At the inference state, we take a WSI, extract patches and construct the graph, and make gene prediction at all spots/patches jointly. 

The overview of our method can be seen in \cref{fig:architecture}. We start by extracting patch-wise image features using a reasonably powerful pre-trained patch encoder, which is currently popular in the digital pathology community (see \cref{sec:encoder} for details).
The next step is to construct a graph that facilitates interactions between biologically relevant spots. We propose both spatial and feature-space clustering to build a multi-faceted hierarchical graph, and incorporate edges within clusters and across clusters. See \cref{sec:graph-construction}.
Finally, we employ a Graph Attention Network (GAT) \cite{velickovic2018graph} that takes in this graph as input and produces an m-dimensional gene expression vector as the output at each node (\cref{sec:gat}), where m is the number of genes. These steps are elaborated in the following subsections.


\subsection{Graph Construction}
\label{sec:graph-construction}
We first describe how to construct the graph from spots of a WSI.
In any ST tissue sample, we would expect spots to have varying degrees of similarity among themselves. These similarities are reflected in the transcriptome where spots have highly correlated gene expressions. To enable our GNN to predict gene expressions more accurately, we need to model the interactions between such similar spots within our graph.

The intuition behind our multi-faceted graph construction is that tissue morphology corresponds to the similarity among spots, and such similarities can occur both within and outside the context of spatial neighborhoods. Firstly, it is natural to expect that spots which are adjacent to one another in the tissue space are likely to have similar tissue morphology. This also means that they are likely to exhibit similar gene expressions. It is, therefore, essential to facilitate interactions among such spots within our graph representation. To accomplish this, we cluster the spots spatially using their tissue position coordinates. 

Secondly, it is important to note that belonging to different spatial clusters does not automatically imply that two spots are different in morphology or gene expression. A tissue sample is likely to have groups of morphologically similar spots that are distant from one another but possess similar gene expressions. But spatial neighborhoods can not capture the interactions among such spots. To model such long-range interactions among spots, we cluster the spots in the feature space using their patch embeddings. The expectation is that spots with similar imaging features are likely to have similar morphology and therefore similar gene expressions. By grouping such spots together, we facilitate efficient message passing and help the GNN make morphology-informed gene expression predictions. The key is that the two clusterings are capturing different aspects of the spot-spot interaction, and we should use both for our graph construction, hence the name ``multi-faceted''.
Once we have clustered the spots based on their tissue position coordinates and patch embeddings, we need to draw the edges while keeping the graph reasonably sparse. We draw two kinds of edges to facilitate intra-cluster and inter-cluster communication. The clustering steps are illustrated in \cref{fig:architecture}(b), followed by graph construction.

\myparagraph{Internal edges.}
For both spatial and feature-space clustering, we first cluster \(n\) spots in a sample into \(c\) clusters. For each of the \(c\) clusters we select a representative spot. The representative spot can either be a random spot in the cluster or be guided by some central tendency. For our method, we use the spot closest to the feature-space centroid of the cluster. This design choice is justified in our ablation studies (see \cref{subsec:ablation_graph}). We call this the \textit{centroid spot}. We connect all spots in a cluster to its centroid spot. We call these edges the \textit{internal edges}. In the spatial clusters, the internal edges enable communication among spatial neighbors. In the feature-space clusters, they enable communication among distant spots of similar morphology. The centroids, as well as, internal edges from both clusterings are shown in \cref{fig:architecture}(b), third and fourth rows. 
Note that the two clusterings will create two sets of internal edges, but the nodes are not duplicated. For a sample consisting of \(n\) spots, each non-centroid spot will be adjacent to two internal edges, connecting it to its spatial cluster centroid and feature cluster centroid respectively. Overall, we have $O(n)$ internal edges.

\myparagraph{Shortcut edges.}
Aside from edges within clusters, we also add edges connecting different clusters to facility long-range communication.
We take the \textit{centroid spot} for all clusters (from both spatial and feature clusterings) and form a complete graph among them. We call these edges \textit{shortcut edges}. Owing to the presence of these shortcut edges and internal edges, \textit{any node in the graph can connect to any other node in a maximum of three hops}. This is crucial in facilitating the GNN to learn long-range interactions effectively. Assuming $c$ clusters for each clustering, we will have ${\binom{2c}{c}} = \frac{2c(2c-1)}{2} = c(2c-1)$ shortcut edges, or fewer if some centroids overlap.

Finally, we will also use the 1-hop spatial proximity edges, connecting each spot with its immediate 8 neighbors.


\subsection{Graph Neural Network}
\label{sec:gat}
We choose Graph Attention Network (GAT) \cite{velickovic2018graph} as our GNN, which enables our model to employ the attention mechanism \cite{vaswani2017attention} on the data. It takes the homogeneous graph as input and outputs a m-dimensional gene expression vector at each node, representing the gene expressions for each spot. The architecture diagram (\cref{fig:architecture}) outlines the four GATConv layers where all layers except the final one employs \(h=8\) attention heads. Layer normalization \cite{ba2016layer} is applied after each GATConv except the last. Additionally, at the beginning of each forward pass, an edge dropout is performed with a probability of \(p=0.2\). For our model, we decided to go with an MSE loss where the optimization objective is to minimize the loss between the ground truth gene expressions and the predictions from the GAT.

\subsection{Patch Encoder}
\label{sec:encoder}
Finally, we discuss the patch encoder used to generate features for each patch. We have experimented with two encoder modules - a ResNet \cite{he2016deep} based patch encoder and the multi-resolution encoder modules from TRIPLEX \cite{chung2024accurate}, which combines features from the patch of interest, neighboring patches, and the whole slide. All of our final experiments use ResNet18 as a backbone. In the pre-training phase, the last layer of the ResNet18 module is discarded and two linear layers project the features into a $d$-dimensional vector. This segment of the encoder was trained on the TCGA-BRCA dataset. After that, we fine tune the encoder to our regression task by adding a linear layer that projects the d-dimensional feature vector into m-dimensional gene expression vector. We use Mean Squared Error (MSE) as our loss. This ensures that the feature extraction step is also gene-aware. We present our findings in the experiments section (see \cref{sec:exp}). After fine-tuning, the projection layer is discarded and the $d$-dimensional feature vector from the penultimate layer is used as the patch embedding, where \(d=256\).
\section{Smoothing Techniques}
\label{sec:smoothing}
Transcriptomics techniques are prone to events where entire subsets of genes are missed during quantification. This results in a gene expression matrix where other than biologically occurring zero expressions, there exists a lot of artificial zeros or \textit{dropouts} \cite{doi:10.1126/science.aaf2403, ding2020systematic}. Past studies have used spatial smoothing on ST data, which leads to an implosion of already sparse data. Additionally, the smoothing does not take gene information into account. We demonstrate here that a two-factor smoothing method is much more robust in smoothing ST data and can help correlate gene expressions with tissue morphology.

\begin{figure}[t]
    \centering
    \includegraphics[width=1\linewidth]{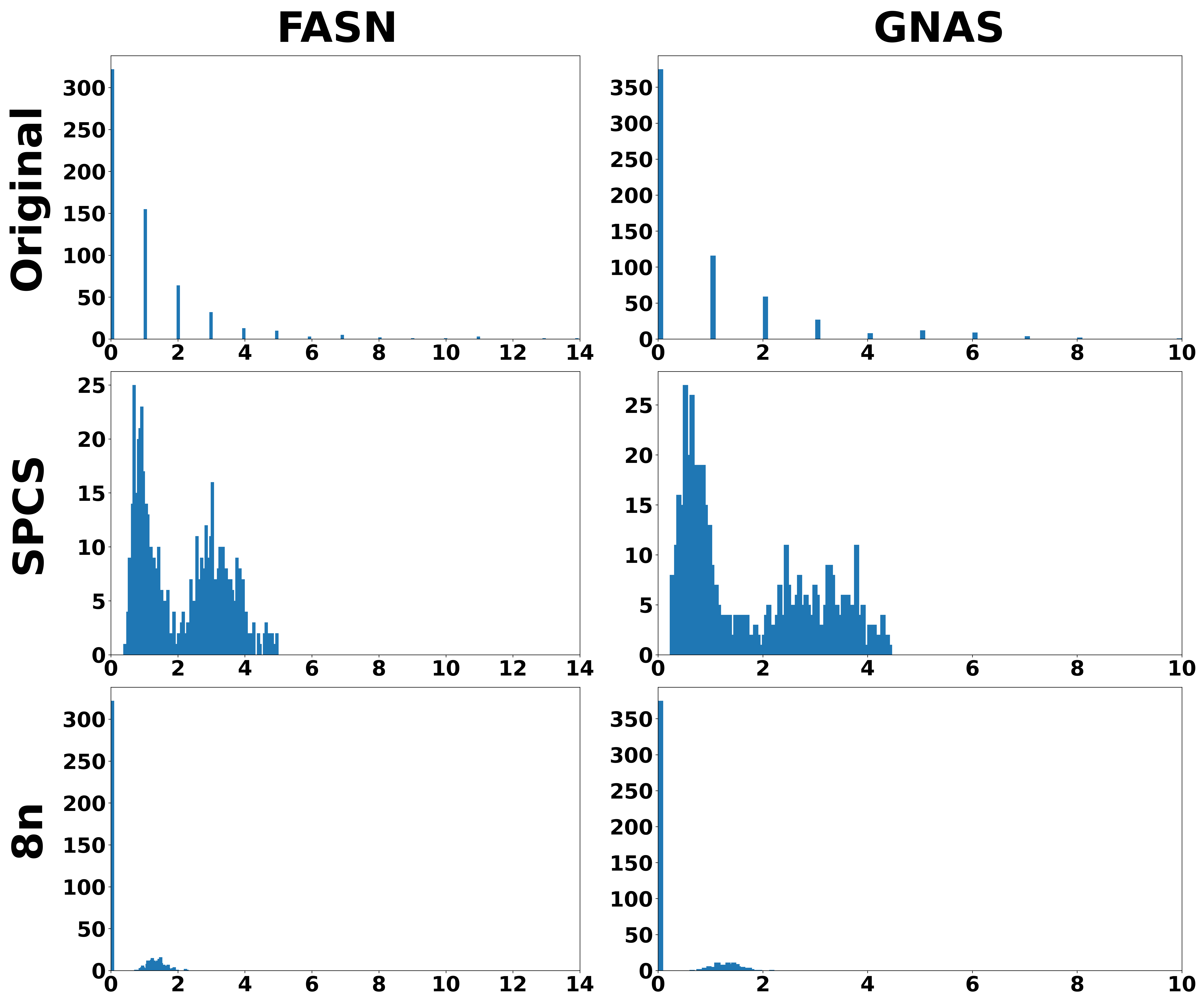}
    \caption{The two columns show the original and smoothed expressions for two cancer-relevant genes (FASN and GNAS) in a tissue sample. The X-axis represents bins of gene expression values, while the Y-axis shows the number of spots belonging to those bins. It is evident that 8n smoothing drastically reduces the scale of the expression values. While SPCS also reduces the scale of the data, it does so being transcriptome-aware.}
    \label{fig:smoothing_histograms}
    \vspace{-0.25in}
\end{figure}

\begin{figure}[t]
    \centering
    \includegraphics[width=1\linewidth]{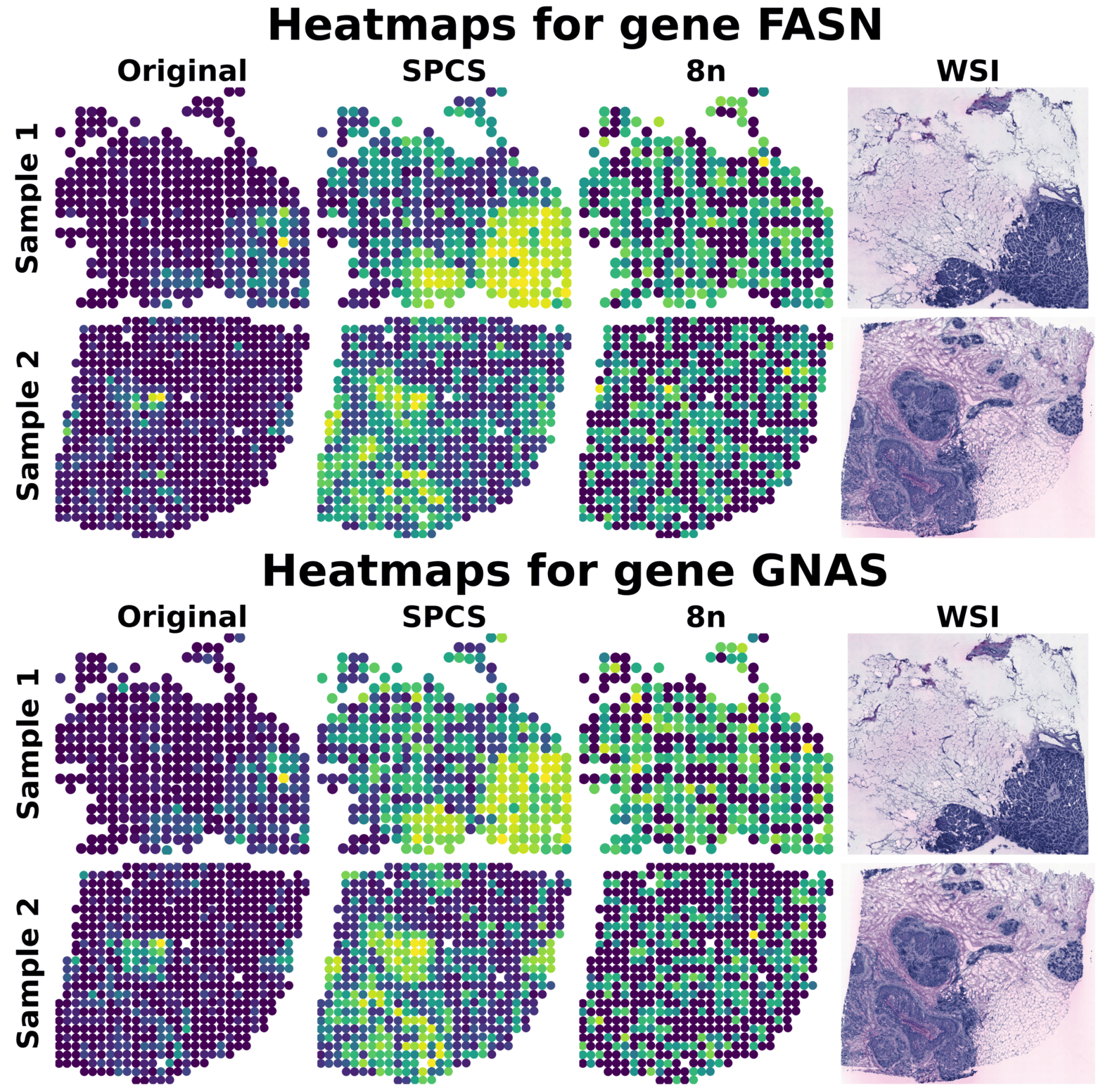}
    \caption{From left to right the four figures in each row represent the original gene expressions, SPCS smoothed gene expressions, 8n smoothed gene expressions, and the WSI. We can see that for both genes and both samples, SPCS outputs have clear correspondence with tissue morphology.}
    \label{fig:smoothing_heatmaps}
    \vspace{-0.25in}
\end{figure}

\begin{figure}
    \centering
    \includegraphics[width=1\linewidth]{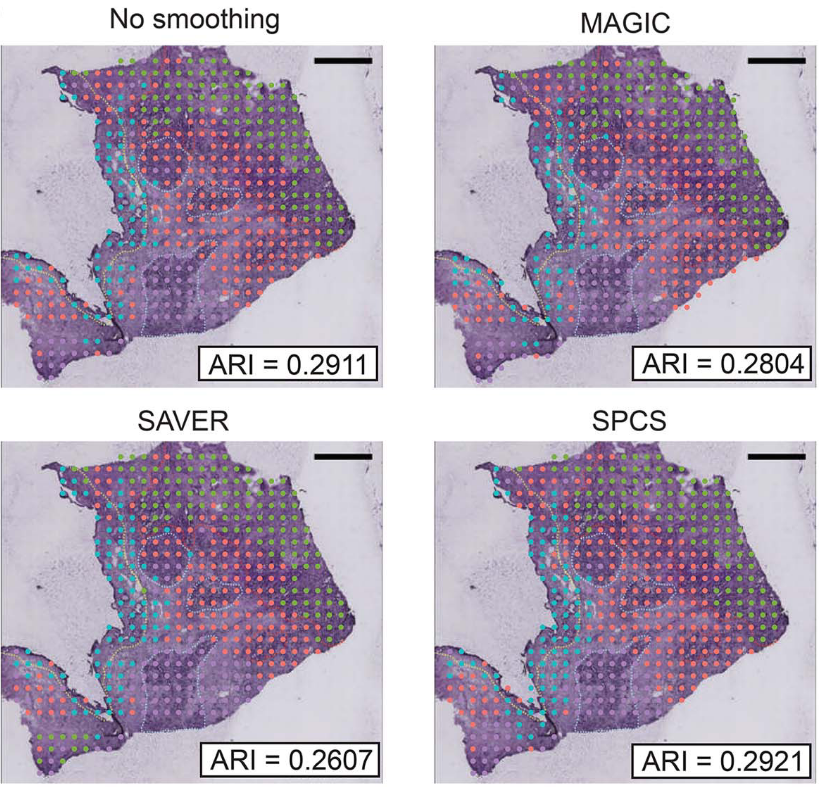}
    \caption{SPCS improves gene expression correspondence with histology labels. Each panel shows the adjusted rand index (ARI) of K-Medoids clusters using various gene expression matrices. Comparison is directly taken from the original SPCS~\cite{10.1093/bib/bbac116} paper.}
    \label{fig:smoothing_ari}
    \vspace{-0.25in}
\end{figure}

\myparagraph{8n Smoothing.}
Past studies on gene expression prediction using ST data \cite{he2020integrating, chung2024accurate} use a form of spatial smoothing which we choose to call \textit{eight-neighbors (8n) smoothing}. In this approach, the gene expression of each spot is replaced with the average of itself and its eight neighbors in the standard square grid structure of ST data. The issue we found with this smoothing approach is that it results in a gene expression matrix which is dramatically reduced in scale. The natural sparsity of the gene expression matrix means that such spatial smoothing pushes down the scale of these values further. While large outliers have a useful impact of lifting up the neighboring values, in the vast majority of the cases, the result is these outliers getting spread across neighboring zeros and their influence getting diffused \cite{doi:10.1126/science.aaf2403, ding2020systematic}. Furthermore, such spatial smoothing will only be spatial association-aware, and not transcriptome-aware.

\myparagraph{SPCS Smoothing.}
In contrast to the one-factor 8n smoothing technique, \textit{SPCS} \cite{10.1093/bib/bbac116} is a two-factor method that performs both spatial and gene expression pattern-based smoothing. SPCS smooths each spot with a weighted contribution from all other spots in a sample. The contribution of each other spot to the smoothing of a target spot consists of two factors:
\begin{itemize}
    \item \textbf{Spatial contribution} is inversely proportional to the Manhattan distance of spot coordinates. Therefore, further spots have rapidly decaying contributions to any target spot.
    \item \textbf{Pattern contribution} is determined by similarity of gene expressions. Initially, a Principal Component Analysis (PCA) \cite{abdi2010principal} is performed to project the gene expressions to a 10D PCA space. The 10D vectors are then used to calculate the Pearson correlation distance (\(1-PCC\)) among spots. An exponential function on these distances yields the pattern contribution.
\end{itemize}

\myparagraph{Comparison.}
\cref{fig:smoothing_ari} outlines the performance comparison between SPCS and two classical smoothing techniques for scRNA-seq data. We can see that the classical techniques do not work well on ST data when compared to SPCS. \cref{fig:smoothing_heatmaps} shows heatmaps of 8n and SPCS outputs alongside the original gene expression values in two samples, for cancer-relevant genes FASN and GNAS. It is evident that the output of SPCS has better alignment with tissue morphology than 8n. The histograms in \cref{fig:smoothing_histograms} show that besides not capturing any morphologically relevant pattern, the output scale of 8n smoothing is very low. On the contrary, while SPCS does downscale the expression values, it still preserves meaningful patterns. See \cref{tab:ablation_smoothing} for ablation results on using SPCS smoothing instead of 8n smoothing.
\section{Experiments}
\label{sec:exp}
Here we discuss the ST data used in our experiments, the pre-processing step, evaluation metrics, implementation details, ablation and smoothing techniques.

\myparagraph{Dataset.}
A standard Spatial Transcriptomics (ST) dataset comprises of histology images of tissue samples, paired with spatially resolved mRNA sequencing output that grounds each spot to a physical location on the tissue. In the original ST data format \cite{doi:10.1126/science.aaf2403}, each tissue sample is sequenced with a spot diameter of 100 \(\mu m\) and a center-to-center diameter of 200 \(\mu m\). ST-Net~\cite{he2020integrating} (68 samples from 23 patients) and Her2ST~\cite{andersson2020spatial} (36 samples from 8 patients) are both breast cancer datasets, while SCC~\cite{ji2020multimodal} (12 samples from 4 patients) is a skin cancer dataset. The average number of in-tissue spots per sample are around 450, 378 and 723 for the three datasets respectively. The TRIPLEX paper uses a subset of 250 most expressed genes for each of the three datasets. We follow suit and use those same gene subsets for all our experiments.

\myparagraph{Pre-processing.}
Patches are extracted from \(\times20\) magnification histology images. Each patch of size \(224\times224\) pixels at \(\times20\) magnification has its centroid at the tissue position coordinates of the spots. We smooth the gene expression data using SPCS \cite{10.1093/bib/bbac116}, a two-factor technique employing a spatial and gene expression pattern-based smoothing. All of the baseline and comparison results in the results section are generated by training and testing the original methods on this smoothed data.

\myparagraph{Evaluation metrics}
We perform cross-validation testing to evaluate model performance both in the pre-training phase for our ResNet18 based encoder and for our GNN training. For all datasets we perform eight-fold cross validation across all slides and report the average results in the results section. The presented results include three metrics - mean squared error (MSE), mean absolute error (MAE) and pearson correlation coefficient (PCC).

\myparagraph{Baselines.}
The cross validation results on two breast cancer datasets - ST-Net~\cite{he2020integrating} and Her2ST~\cite{andersson2020spatial} - as well as a skin cancer dataset - SCC~\cite{ji2020multimodal} - compares performances with six baselines. The baselines are - a ResNet18 encoder with an FCN for gene expression prediction, BLEEP, HisToGene, Hist2ST, THItoGene~\cite{jia2024thitogene} and TRIPLEX. For fair comparison we use the same ResNet18 architecture as our own patch encoder in BLEEP and TRIPLEX as well, and train all of these models on the SPCS smoothed data.

\begin{table*}[t]
\small
\centering

\resizebox{\linewidth}{!} {%
    \begin{tabular}{|c|c|c|c|c|c|c|c|c|c|c|}
        \hline
        \multirow{2}{*}{\textbf{Method}} & \multirow{2}{*}{\textbf{Graph}} & \multicolumn{3}{|c|}{\textbf{ST-Net}} & \multicolumn{3}{|c|}{\textbf{Her2ST}} & \multicolumn{3}{|c|}{\textbf{SCC}} \\
        \cline{3-11}
        & & \textbf{MSE \(\downarrow\)} & \textbf{MAE \(\downarrow\)} & \textbf{PCC \(\uparrow\)} & \textbf{MSE \(\downarrow\)} & \textbf{MAE \(\downarrow\)} & \textbf{PCC \(\uparrow\)} & \textbf{MSE \(\downarrow\)} & \textbf{MAE \(\downarrow\)} & \textbf{PCC \(\uparrow\)} \\
        \hline
        ResNet+FCN & - & \(0.1999 \) & \(0.3448\) & \(0.5221 \) & \(0.6623 \) & \(0.6385\) & \(0.4629 \) & \(0.6103 \) & \(0.6290\) & \(0.4619 \)  \\
        BLEEP & - & \(0.3756 \) & \(0.4736\) & \(0.0784 \) & \(0.7426 \) & \(0.6591\) & \(0.2747 \) & \(0.6079 \) & \(0.6013\) & \(0.4176 \) \\
        HisToGene & - & \(0.3054 \) & \(0.4336\) & \(0.1211 \) & \(0.9452 \) & \(0.7739\) & \(0.2062 \) & \(\mathbf{0.3095} \) & \(\mathbf{0.4367}\) & \(0.1225 \) \\
        Hist2ST & \textit{1h4} & \(0.3811 \) & \(0.4822\) & \(0.1525 \) & \(0.7843 \) & \(0.7286\) & \(0.2479 \) & \(1.0190 \) & \(0.7639\) & \(0.3003 \) \\
        THItoGene & \textit{1h4} & \(0.2925\) & \(0.4111\) & \(0.3666\) & \(0.8436 \) & \(0.7069\) & \(0.3445\) & \(0.6798\) & \(0.6442\) & \(0.3897\) \\
        TRIPLEX & - & \(0.1472 \) & \(0.2943\) & \(0.2320 \) & \(0.8982 \) & \(0.6946\) & \(0.3927 \) & \(0.4891\) & \(0.5356\) & \(0.5416\) \\
        \hline
        \multirow{2}{*}{\textbf{MERGE}} & \textit{1h*} & \(0.1783 \) & \(0.3207\) & \(0.5789 \) & \(0.6531 \) & \(0.6270\) & \(0.4866 \) & \(0.5706 \) & \(0.6029\) & \(0.4843 \) \\
         & \textit{hier} & \(\mathbf{0.1347 }\) & \(\mathbf{0.2834}\) & \(\mathbf{0.6795 }\) & \(\mathbf{0.6422 }\) & \(\mathbf{0.6255}\) & \(\mathbf{0.5037 }\) & \(0.5353 \) & \(0.5838\) & \(\mathbf{0.5512} \) \\
        \hline
    \end{tabular}
}
\vspace{-0.1in}
    \caption{Cross-validation performance on the ST-Net, Her2ST and SCC datasets including baselines. (\textit{1h4} and \textit{1h*} denote graphs of four and all 1-hop neighbors respectively, while \textit{hier} indicates multi-faceted hierarchical graph construction.)}
    \vspace{-0.25in}
    \label{tab:cross_val}
\medskip
\end{table*}

\subsection{Quantitative Results}
\myparagraph{Cross-validation Performance.} From \cref{tab:cross_val}, we can see that using graph neural networks improves correlation in general, and it improves the other metrics when used with our ResNet18 based encoder. We can see that upon using GNN with simple 1-hop edges, the improvement in PCC is 0.057 when the encoder backbone is ResNet18. Furthermore, there is an additional improvement upon incorporating the multi-faceted hierarchical graph construction strategy. It improves only marginally by 0.011 when using the multi-resolution encoder, but improves by a more substantial margin of 0.101 when using the ResNet18 based encoder. We can see that when we use a ResNet18-based encoder instead of the multi-resolution encoder, we observe slight improvements across the board, with MSE, MAE, and PCC improving by 0.011, 0.013, and 0.023. This indicates that the bulk of the performance gap can be traced back to the use of GNNs, while a smaller fraction of it can be attributed to the graph construction strategy.



\subsection{Qualitative Results}
\begin{figure}
    \centering
    \includegraphics[width=1\linewidth]{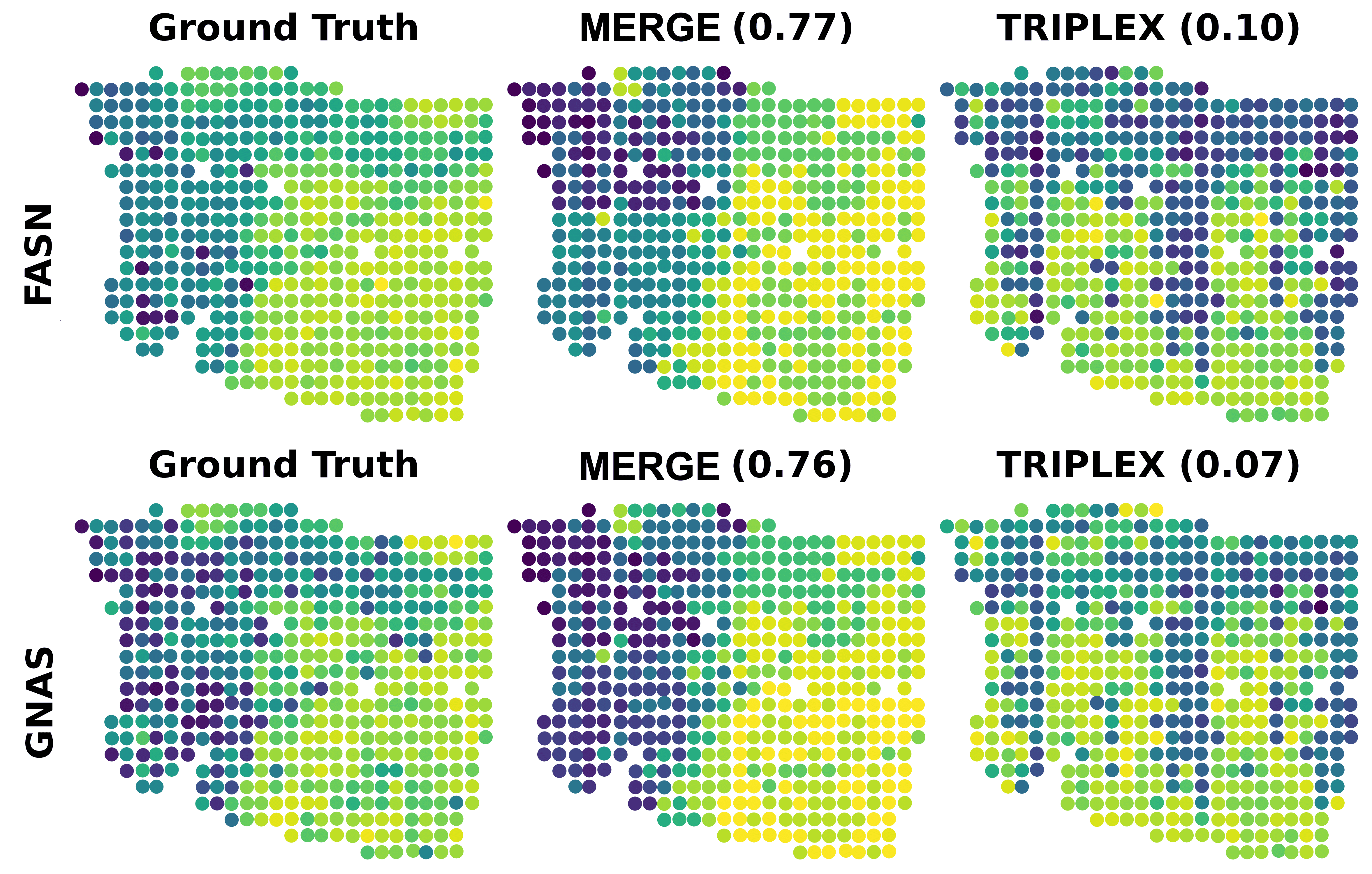}
    \caption{The figure shows the predictions for two cancer-relevant genes (FASN and GNAS) as heatmaps on the tissue space, with the PCC in parentheses. The plots show that our method achieves higher correspondence with ground truth expressions.}
    \label{fig:heatmaps}
\end{figure}

\myparagraph{Predictions.}
We plot the gene expressions for known relevant marker genes for breast carcinoma on the tissue space and compare our predictions with those of TRIPLEX in \cref{fig:heatmaps}. The gene expression heat-maps are shown for two well known genes - GNAS as a breast cancer biomarker \cite{jin2019elevated} and FASN as a tumor biomarker \cite{hunt2007mrna}. It is evident that MERGE achieves significantly higher correlations (0.77 and 0.76) for both genes.

\begin{figure}
    \centering
    \includegraphics[width=1\linewidth]{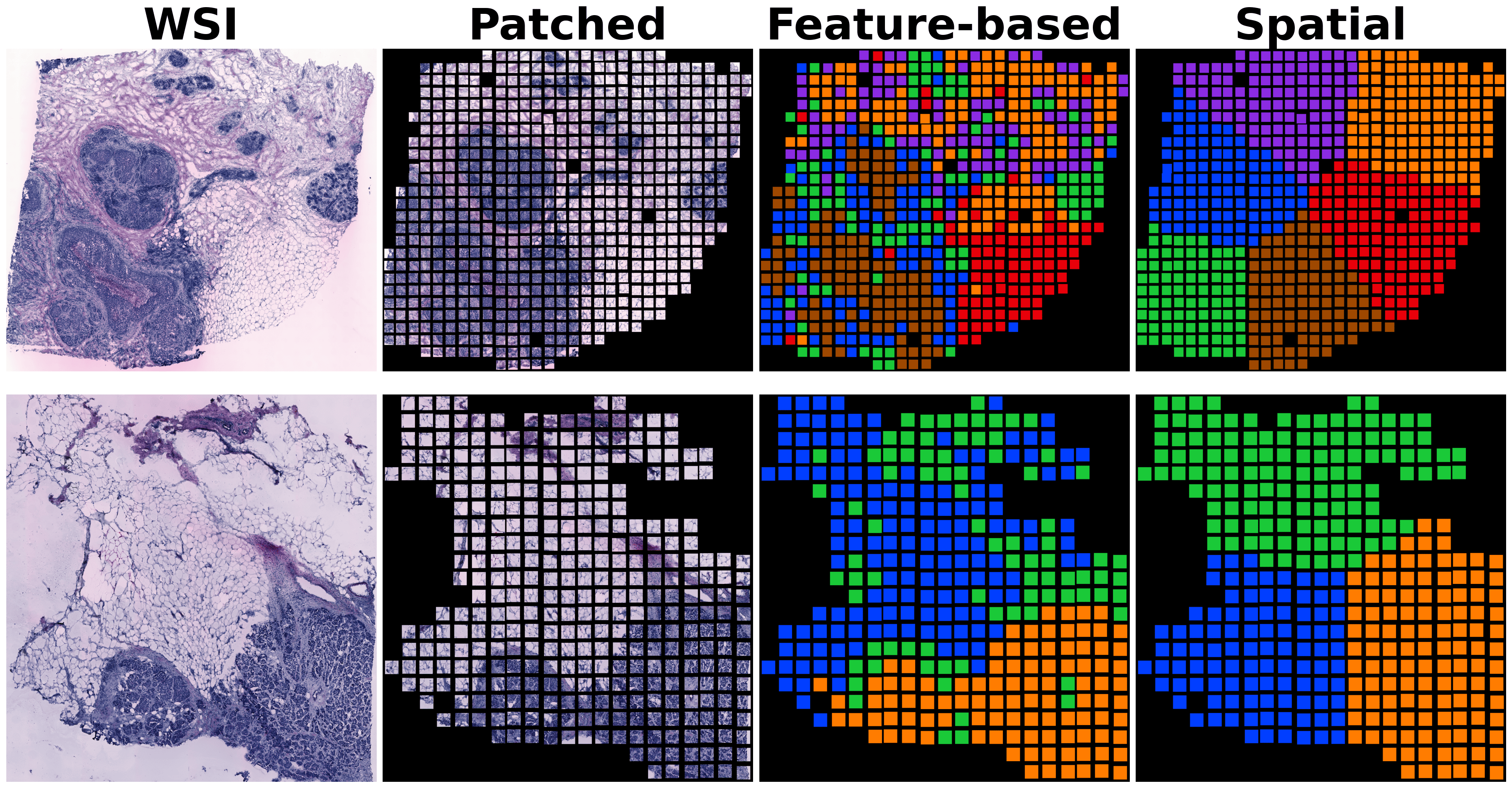}
    \caption{The figure shows clustering outputs for two samples in two rows. From left to right we see the WSI, extracted patches, feature space clusters and spatial clusters.}
    \label{fig:cluster_viz}
    \vspace{-0.25in}
\end{figure}

\myparagraph{Multi-faceted Hierarchical Graph.}
The hierarchical clustering strategy we employ captures better gene correlation. The cluster visualizations shown in \cref{fig:cluster_viz} shows the outputs of the disjoint clustering steps. We can see that the feature-space clustering captures tissue morphology quite effectively. The selection of centroid spots to represent each cluster ensures that the graph remains reasonably sparse. This inevitably helps the GNN learn better representations while managing computational intensity.

\begin{table}
    \centering
    \begin{tabular}{|c|c|c|c|}
        \hline
        & \textbf{MSE} & \textbf{MAE \(\downarrow\)} & \textbf{PCC \(\uparrow\)} \\
        \hline
        1-hop & \(0.1783\) & \(0.3207\) & \(0.5789\) \\
        \hline
        w/o spatial & \(0.1539\) & \(0.3017\) & \(0.6356\) \\
        \hline
        w/o feature & \(0.1377\) & \(0.2879\) & \(0.6719\) \\
        \hline
        random centroid & \(0.14\) & \(0.2897\) & \(0.666\) \\
        \hline
        spatial centroid & \(0.1431\) & \(0.2921\) & \(0.6583\) \\
        \hline
        \textbf{MERGE} & \(\mathbf{0.1347}\) & \(\mathbf{0.2834}\) & \(\mathbf{0.6795}\) \\
        \hline
    \end{tabular}
    \caption{Using spatial or feature space clustering as the sole clustering method improves results, but the combined method yields the best results. Additionally, using the spatial centroid instead of the feature centroid as the centroid node yields poorer results.}
    \label{tab:ablation_graph}
\end{table}

\begin{table}
    \centering
    \begin{tabular}{|c|c|c|c|}
        \hline
        \textbf{Cluster size} & \textbf{MSE} & \textbf{MAE \(\downarrow\)} & \textbf{PCC \(\uparrow\)} \\
        \hline
        25 & \(0.1367\) & \(0.2845\) & \(0.6754\) \\
        \hline
        75 & \(0.1355\) & \(0.284\) & \(0.6768\) \\
        \hline
        \textbf{100} & \(\mathbf{0.1347}\) & \(\mathbf{0.2834}\) & \(\mathbf{0.6795}\) \\
        \hline
        150 & \(0.1372\) & \(0.2866\) & \(0.673\) \\
        \hline
        200 & \(0.1377\) & \(0.2864\) & \(0.674\) \\
        \hline
    \end{tabular}
    \caption{Increasing the cluster size improves performance to a point, but beyond a certain cluster size, clusters are not morphologically meaningful anymore. We see a drop in performance beyond a cluster size of 100, for the ST-Net dataset.}
    \label{tab:ablation_cluster_size}
    \vspace{-0.25in}
\end{table}

\subsection{Ablation}
\label{subsec:ablation_graph}
This section outlines the contribution of a few core design choices and components. We perform ablation studies to evaluate model performance with and without our multi-faceted hierarchical graph construction strategy. We also assess the utility of selecting the spot closest to the feature-centroid as the representative for each cluster, alongside the impact of cluster size on performance. Finally we test the efficacy of SPCS smoothing. All of these studies are performed on the ST-Net dataset.

\myparagraph{Graph Construction.}
With all other parameters being the same, this ablation study evaluates using the two disjoint clustering steps, as well as the multi-faceted graph construction strategy. When comparing the standalone effectiveness of using only spatial or feature-space clustering with the combined method, we find that the combined method works better (see \cref{tab:ablation_graph}). Another design choice was the selection of the representative node for each cluster. We ran experiments with the representative node being any random node, the node closest to the spatial centroid, and the node closest to the feature centroid. \cref{tab:ablation_graph} shows that using the feature centroid yields the best results.

\myparagraph{Cluster Size.}
The number of spots within a cluster can affect model performance to a certain degree. The cluster size is therefore a tunable hyperparameter. Our experiments with various cluster sizes shows that increasing the cluster size helps improve performance, but only to a certain degree. Beyond a certain point, the clusters become too large to isolate morphologically distinct segments. \cref{tab:ablation_cluster_size} shows that having a cluster size larger than 100 starts to hurt performance. This intuitively makes sense as the average number of spots per sample in the ST-Net dataset is 450. Having more than 100 spots per cluster results in fewer than four clusters for most samples and the clusters fail to separate morphologically distinct spots.

\myparagraph{SPCS Smoothing.}
To understand the contribution of the smoothing method in model performance, we also trained two baselines as well as our final models with either patch encoders on both smoothing methods. Except in the case of the original TRIPLEX method, the other methods exhibit lower MSE and higher PCC using SPCS smoothed data (see \cref{tab:ablation_smoothing})

\begin{table}
    \centering
    \begin{tabular}{|c|c|c|c|}
         \hline
         \textbf{Method} & \textbf{Smooth} & \textbf{MSE \(\downarrow\)} & \textbf{PCC \(\uparrow\)} \\
         \hline
         \multirow{2}{*}{ResNetSimCLR} & 8n & \(0.3349\) & \(0.2701\) \\
         & SPCS & \(0.1999\) & \(0.5221\) \\
         \hline
         \multirow{2}{*}{TRIPLEX} & 8n & \(0.0760\) & \(0.3014\) \\ 
         & SPCS & \(0.1472\) & \(0.2320\) \\
         \hline
         \multirow{2}{*}{MERGE (w/ ResNet)} & 8n & \(0.1387\) & \(0.6645\) \\
         & SPCS & \(0.1355\) & \(0.679\) \\
         \hline
    \end{tabular}
    \caption{Using SPCS smoothing yields better results over using 8n smoothing (except with TRIPLEX).}
    \label{tab:ablation_smoothing}
    \vspace{-0.25in}
\end{table}
\section{Conclusion}
In this paper we present MERGE, an innovative approach for predicting spatial gene expressions from tissue images. MERGE incorporates a multi-faceted hierarchical graph construction strategy, which is grounded in spatial and image feature space clustering. Intra- and inter-cluster edges allow MERGE to model short and long-range interactions among similar tissue segments. Thus, using a GNN setup allows accurate joint gene expression prediction. MERGE outperforms existing methods in standardized metrics, predicting robust, morphology-guided gene expressions in a spatial context.

{
    \small
    \bibliographystyle{ieeenat_fullname}
    \bibliography{main}
}
\clearpage

\setcounter{page}{1}
\maketitlesupplementary

\setcounter{section}{6}
\setcounter{figure}{7}
\setcounter{table}{4}


In the supplementary material, we begin with implementation details in \cref{sec:sup_implementation}. \cref{subsec:sup_exp_setup} discusses the experimental setup, including the hardware information and environment specifications. In \cref{subsec:sup_smoothing_det} we discuss the nuances of the SPCS smoothing method and the details pertaining to its implementation for our purposes. We analyze the influence of cluster size on the feature space clustering and how well it aligns with the gene expressions in \cref{subsec:sup_cluster_size}. \cref{subsec:sup_baselines} discusses the implementation details of two baselines and also analyzes the performance gap in case of one of them. In the second section (\cref{sec:sup_res_analysis}), we further analyze the results of our experiments and present visualizations to that effect. \cref{subsec:sup_pcc_comp} presents a comparison of the PCC scores obtained by MERGE and TRIPLEX~\cite{chung2024accurate} across the ST-Net dataset, while \cref{subsec:sup_ablation_viz} presents visualizations for ablation of the graph construction strategy and its various modules.

\section{Implementation Details}
\label{sec:sup_implementation}
\subsection{Experimental Setup}
\label{subsec:sup_exp_setup}
The ResNet18-based patch encoder is implemented in PyTorch (version 2.2.2). The graph neural network is implemented using PyTorch Geometric (version 2.5.2). Both models are trained on a NVIDIA RTX A6000 GPU. To ensure reproducibility a constant seed (3927) is set across all implementations and reruns of the model. The training of the ResNet18 is performed over 15 epochs, while the GNN is trained over 400 epochs. Each training process is replicated for five times and the best model is picked for each experiment.

\subsection{Smoothing}
\label{subsec:sup_smoothing_det}

\begin{figure}
    \centering
    \includegraphics[width=1\linewidth]{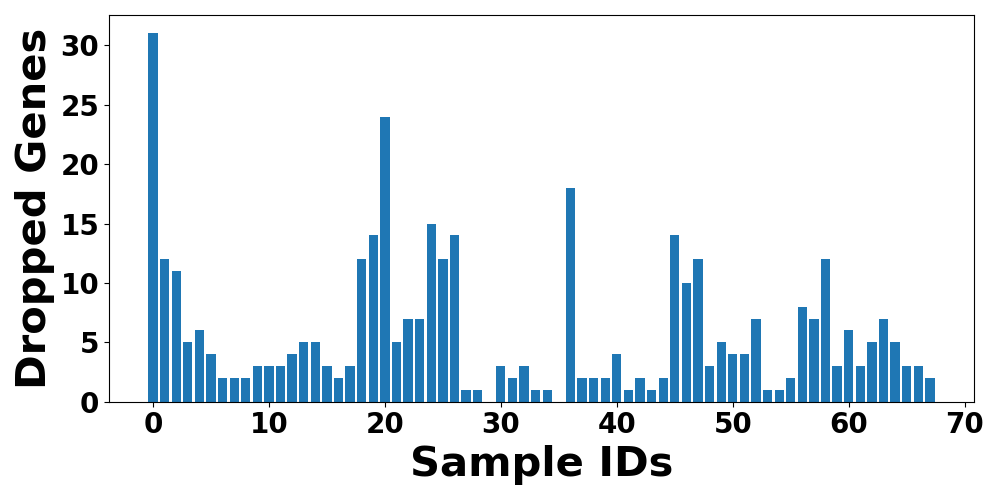}
    \caption{This presents a bar chart of the sample-by-sample dropping of genes based on amount of zero expression in the ST-Net dataset. The average number of genes dropped per sample is \(5.72\). We skip this step and use the 250-gene subset from the original study \cite{he2020integrating})}
    \label{fig:sup_gene_filter}
\end{figure}

\begin{figure*}
    \centering
    \includegraphics[width=1\textwidth]{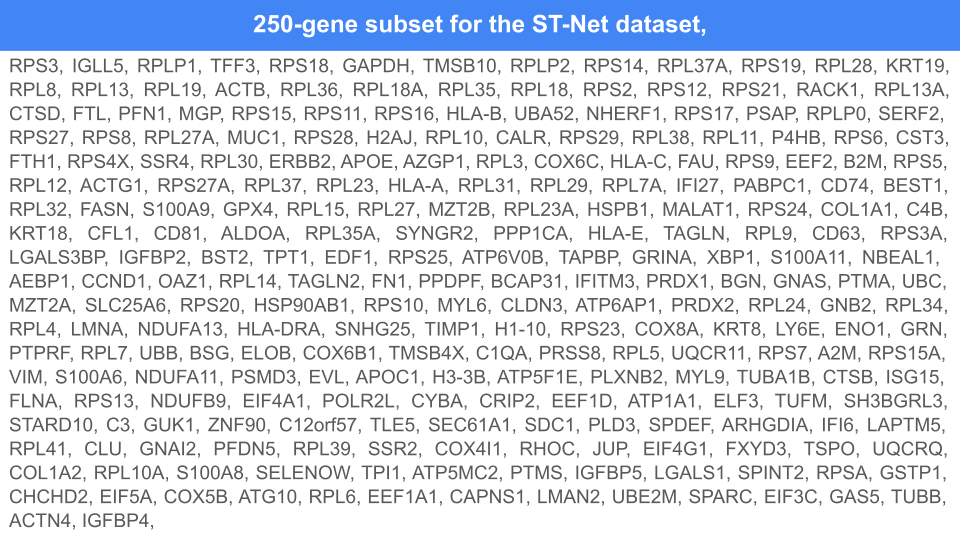}
    \caption{List of the 250 genes used in all experiments on the ST-Net dataset.}
    \label{fig:sup_stnet_genes}
\end{figure*}

\begin{figure*}
    \centering
    \includegraphics[width=1\textwidth]{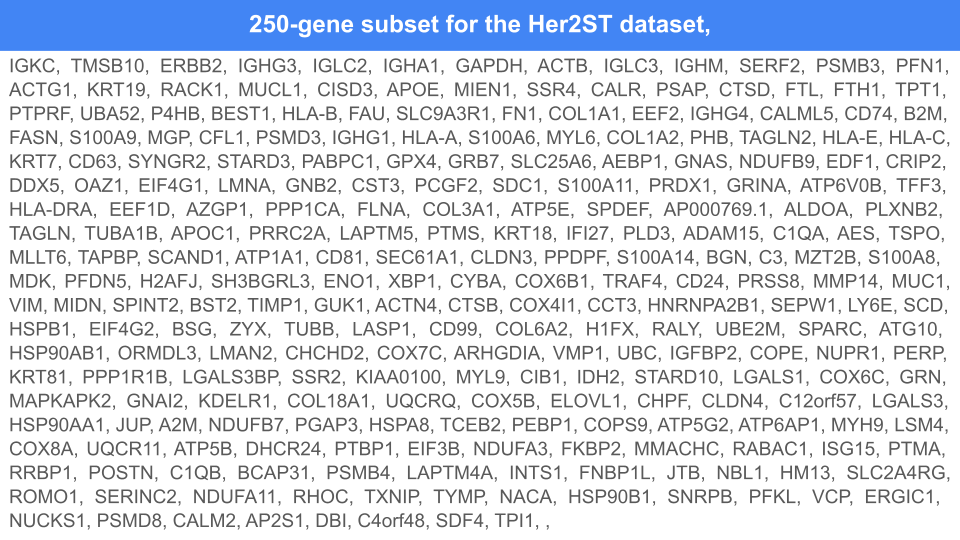}
    \caption{List of the 250 genes used in all experiments on the Her2ST dataset.}
    \label{fig:sup_her2st_genes}
\end{figure*}

\begin{figure*}
    \centering
    \includegraphics[width=1\textwidth]{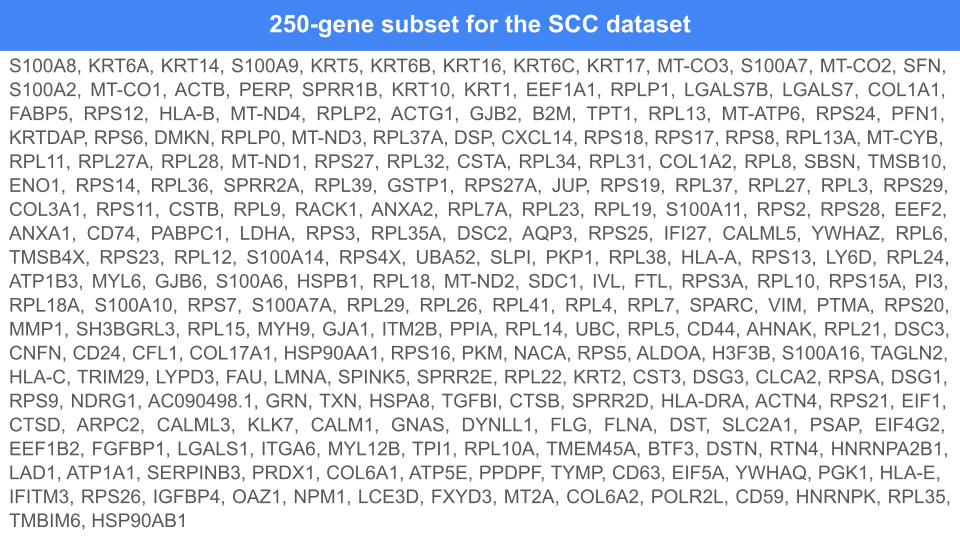}
    \caption{List of the 250 genes used in all experiments on the SCC dataset.}
    \label{fig:sup_scc_genes}
\end{figure*}

We already established the effectiveness of SPCS \cite{10.1093/bib/bbac116} smoothing over the spatial smoothing employed in prior studies \cite{he2020integrating, chung2024accurate}. This section presents further details on the implementation of the SPCS smoothing employed on the ST-Net dataset. The source code for SPCS is obtained from the GitHub repository provided in the original publication. The R-package is compiled and run in a Python kernel. The raw UMI counts and tissue position coordinates for the spots must first be converted to the appropriate data format to be processed by the SPCS code. In the original SPCS implementation, a quality check is performed across the gene library to filter out low-quality genes using a threshold for the percentage of spots in which a gene is expressed as well as the variance of the gene expressions. Typically the zero cutoff parameter is set to \(0.7\), meaning that genes that are zero in over 70\% of the spots are discarded, while the rest are retained. Running this gene filtering step on all samples of the ST-Net dataset reveals that the vast majority of slides retain most of the genes (see \cref{fig:sup_gene_filter}). The average number of genes dropped per sample is \(5.72\). We skip this quality check step for all three datasets, and perform SPCS 250-gene subsets used in the TRIPLEX paper. These genes are listed in~\cref{fig:sup_stnet_genes}, ~\cref{fig:sup_her2st_genes}, ~\cref{fig:sup_scc_genes}. The values for the other parameters in the SPCS method are outlined in \cref{tab:sup_spcs_params}.

\begin{table}[t]
    \centering
    \begin{tabular}{|c|c|}
        \hline
        \textbf{Parameter} & \textbf{Value} \\
        \hline
        Gene Zero Cutoff & N/A \\
        Gene Variance Cutoff & N/A \\
        \(\tau_s\) & 2 \\
        \(\tau_p\) & 16 \\
        \(alpha\) & 0.6 \\
        \(beta\) & 0.4 \\
        Filling threshold & 0.5 \\
        is.hexa & False \\
        is.padding & False \\
        \hline
    \end{tabular}
    \caption{Parameters for running SPCS on the ST-Net dataset.}
    \label{tab:sup_spcs_params}
\end{table}

\subsection{Cluster Size}
\label{subsec:sup_cluster_size}
\begin{figure*}
    \centering
    \includegraphics[width=0.9\textwidth]{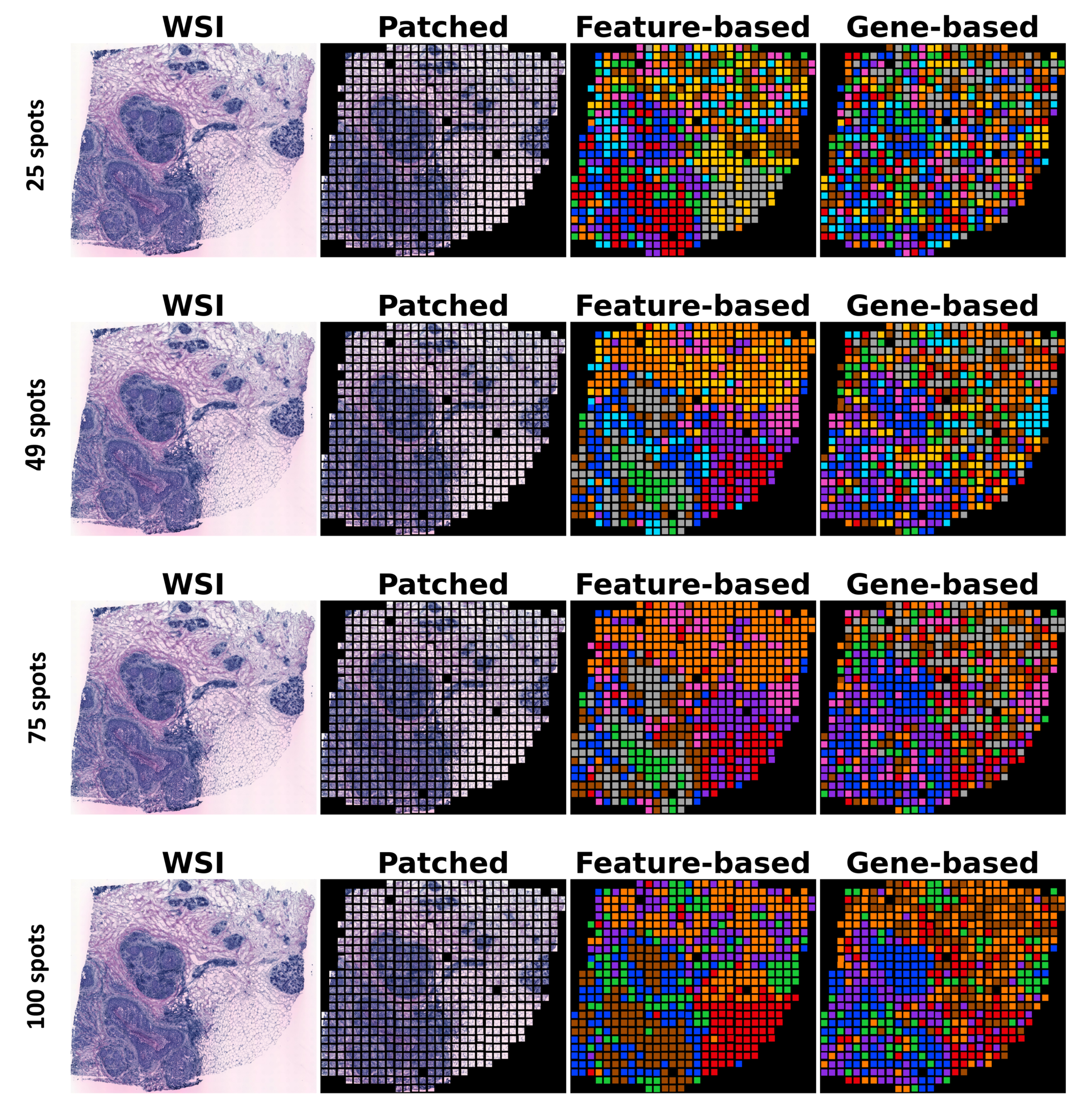}
    \caption{The figure shows the WSI, extracted patches, outputs of feature space clustering, and outputs of gene space clustering using five cancer biomarkers. Each row represents a different cluster size. It is evident that clusters that are too small fail to represent both the morphologically similar tissue segments and the gene space clusters. Larger cluster sizes are more effective at capturing both of these, but still can not accurately align with the gene space clusters. This underlines the necessity of feature space clustering but also depicts why it is not sufficient by itself.}
    \label{fig:sup_clustering_viz}
\end{figure*}

\myparagraph{Spatial Clustering.}
The cluster size impacts the two clustering approaches in different ways. For the spatial clustering, using very small cluster sizes will mean that a smaller number of adjacent spots will be grouped together. This will result in the graph trying to model smaller tissue segments. This can have varying effects depending on the sample. If a sample consists of large homogeneous tissue segments, smaller spatial clusters will not be too helpful in capturing the interactions of spots within them. But if there are small homogeneous tissue segments, using smaller clusters will help the GNN learn from these smaller groups of spots. This will result in a more accurate modeling of the morphology driven interactions among spots, and thereby enhance gene expression prediction.

\myparagraph{Feature Space Clustering.}
In the feature space, the goal is to group spots based on imaging features. The expectation is that morphologically similar spots will have similar imaging features, and therefore, be grouped together. This extends to the idea that morphologically similar tissue segments are likely to exhibit similar gene expressions. Therefore, we should expect to see some alignment among tissue morphology, feature space clusters and the gene expressions. To visualize this we perform clustering in the gene space by using the gene expression vectors as the feature vectors in a clustering scheme similar to feature space clustering. The genes used for this clustering are - FASN, GNAS, XBP1, AEBP1, SPARC, and BGN~\cite{he2020integrating}. These are all known cancer biomarkers. We use the same cluster size for both. \cref{fig:sup_clustering_viz} shows the outputs of feature space clustering and gene-based clustering for various cluster sizes. We can see that smaller clusters fail to capture morphologically meaningful groups well. They also fail to align well with gene space clusters. But larger cluster sizes result in more gene-aligned feature space clusters. We can see that there is still a significant misalignment among the clusters in the image feature space and the clusters in the gene expression space. This is why feature space clustering is not sufficient on its own, and provides better outputs only when combined with spatial clustering.

\subsection{Baselines}
\label{subsec:sup_baselines}

\subsubsection{ResNet+FCN}
The first baseline is composed of our ResNet18 patch encoder followed by a fully connected layer (FCN) to predict the 250 genes per patch. This is the simplest variant of architecture, which is directly inspired by the original ST-Net~\cite{he2020integrating} architecture where a DenseNet followed by a fully connected layer was used for gene expression prediction.

\subsubsection{BLEEP}

\begin{figure}
    \centering
    \includegraphics[width=1\linewidth]{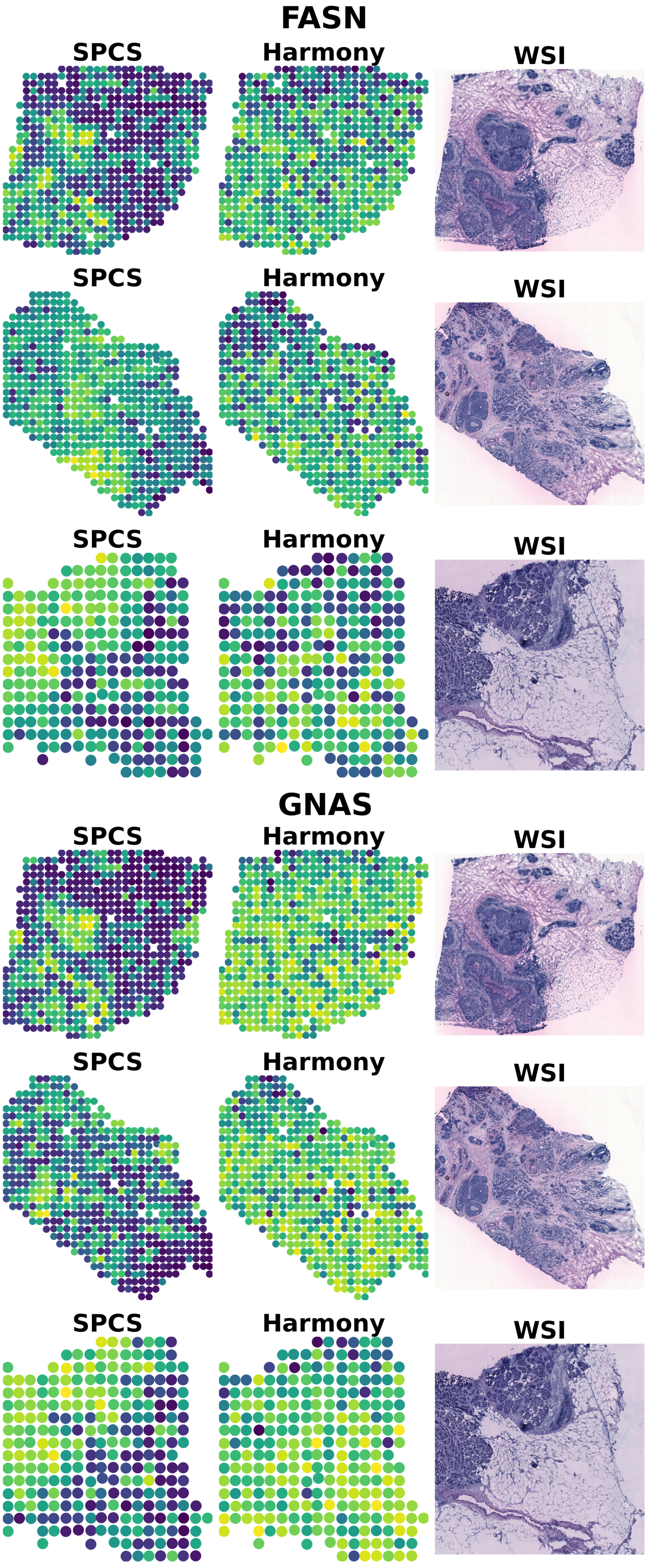}
    \caption{When we plot the expression values for two cancer biomarker genes - FASN and GNAS - on the tissue space, we see that the morphological patterns captured by SPCS are lost upon applying Harmony. This is most likely the cause of the poor performance of BLEEP on the SPCS smoothed ST-Net dataset.}
    \label{fig:sup_harmony_viz}
\end{figure}

The source code of BLEEP~\cite{xie2023spatially} is obtained from the original publication. The same ResNet18 architecture is used here as the patch encoders in both MERGE and TRIPLEX. BLEEP seems to perform rather poorly when trained and evaluated on the SPCS smoothed data. Our assumption is that this is caused by the use of Harmony~\cite{korsunsky2019fast} by BLEEP. Harmony is a batch correction algorithm designed for single cell RNA-seq data (scRNA-seq). Harmony models and removes the artifacts generated by known sources of variation in scRNA-seq data. In applying Harmony to ST data however, each spot must be treated as a cell and this does not translate well in all cases. In case of BLEEP, using Harmony on the raw gene expression data works well with their original four-sample dataset. However, using it on the SPCS smoothed ST-Net data interferes with the gene expression patterns captured by SPCS. We assume that this is what most likely results in a poor performance on the ST-Net dataset. The detrimental effect of applying Harmony on ST-Net data is visualized in \cref{fig:sup_harmony_viz} where we can see that the morphological patterns captured by SPCS for two cancer biomarker genes - FASN and GNAS - are lost in the outputs of Harmony.

\subsubsection{HisToGene, Hist2ST and THItoGene}
The source code of HisToGene~\cite{Pang2021.11.28.470212}, Hist2ST~\cite{10.1093/bib/bbac297} and THItoGene~\cite{jia2024thitogene} are adopted from the original publications. We train and test on SPCS smoothed data for all three baselines. All three models perform logCPM normalization on the gene expression matrix. We remove this portion of the code before training on SPCS data as we already perform logCPM normalization during SPCS smoothing. Hist2ST and THItoGene use 4-nearest-neighbors graphs for their purposes and we adapt that to the other datasets as well. We set the global seed for all modules and libraries the same value as MERGE (3927) for consistency.

\subsubsection{TRIPLEX}
The source code of TRIPLEX~\cite{chung2024accurate} is obtained from the original publication as well. This is used to train and test on both 8n and SPCS smoothed data. Since TRIPLEX performs the spatial smoothing (8n) within the model code, we remove that portion of code as well as the log normalization step while training on the SPCS smoothed data. The remaining parameters are kept unchanged in the original implementation and the parameter specifications are obtained from the supplementary materials provided by the authors. The seed set by TRIPLEX is also updated to match the seed in MERGE (3927) for consistency.

\section{Results Analysis}
\label{sec:sup_res_analysis}
\subsection{Results Comparison - PCC}
\label{subsec:sup_pcc_comp}
\begin{figure*}[t]
    \centering
    \includegraphics[width=1\textwidth]{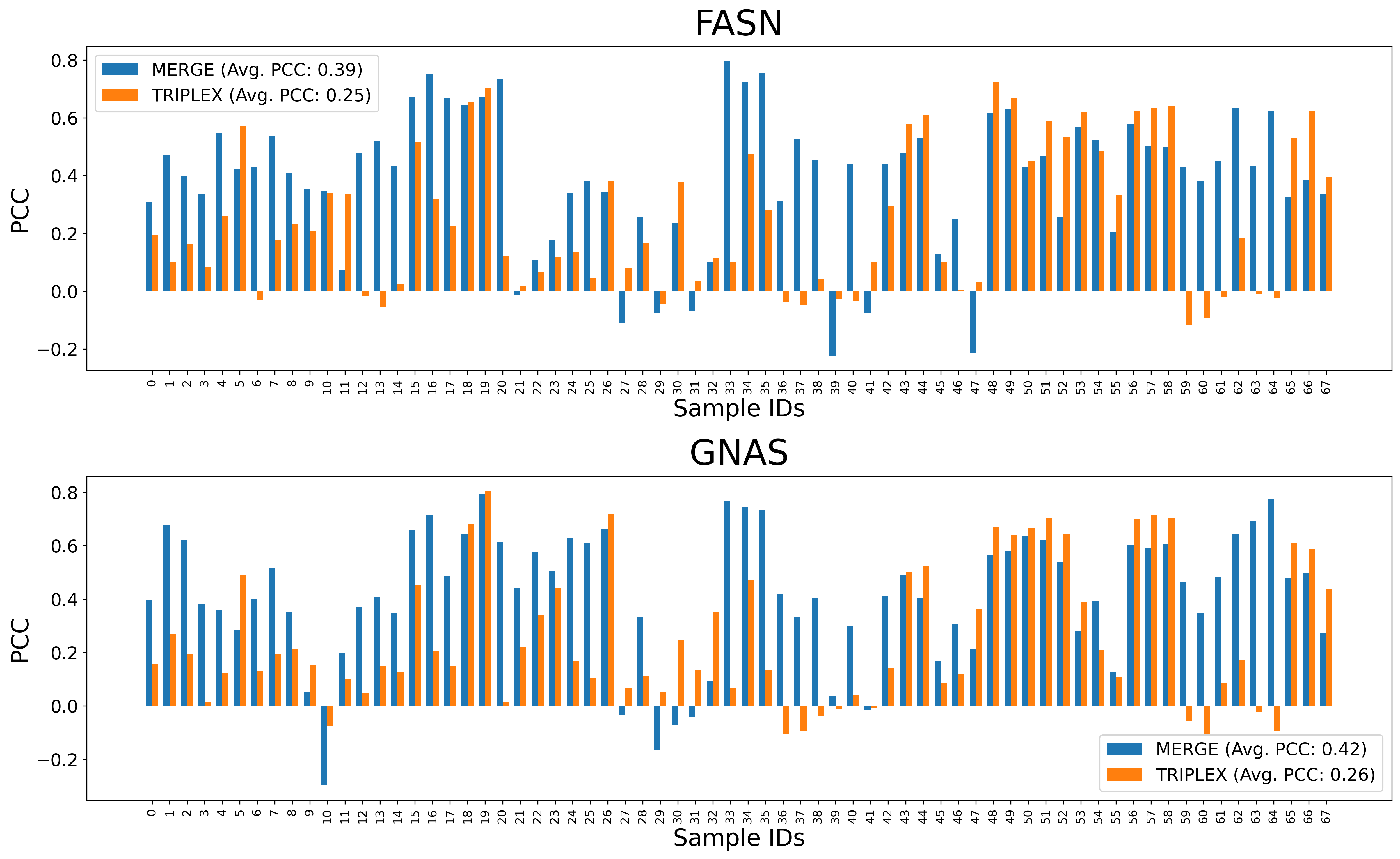}
    \caption{The PCC between the ground truth and predicted gene expressions for the tumor marker FNAS gene and breast cancer biomarker GNAS gene. The average PCC across the dataset is reported inside the legend within parentheses.}
    \label{fig:sup_correlations}
\end{figure*}

\begin{figure*}[t]
    \centering
    \includegraphics[width=1\textwidth]{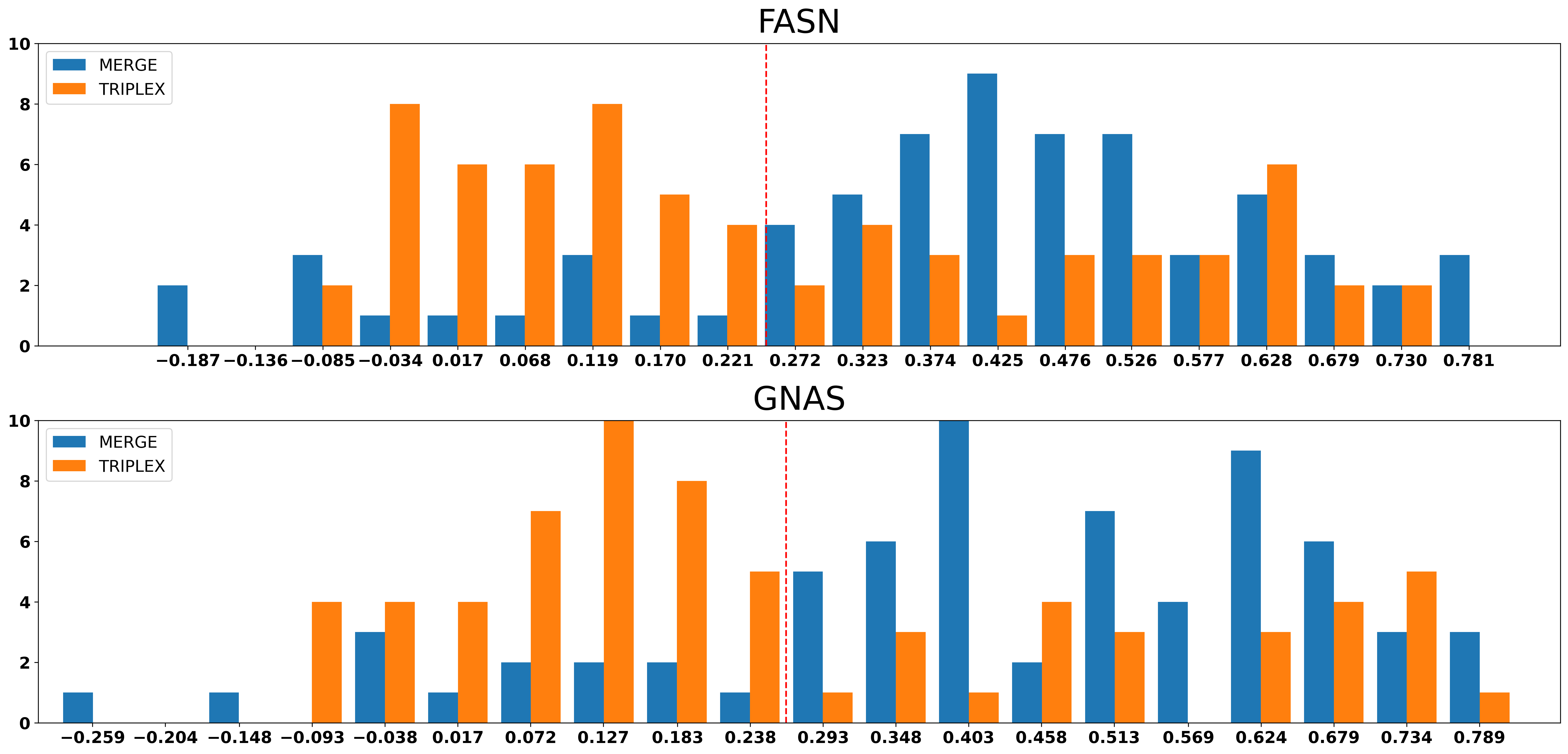}
    \caption{This is a histogram of PCC for the two methods for the ST-Net dataset. The upper panel shows the histogram for the FASN gene, while the lower panel shows the same for the GNAS gene. We can see that MERGE attains low PCC for significantly fewer samples. Additionally, there are very few samples where both methods perform poorly and attain negative PCC for either gene, although this number is slightly higher for MERGE.}
    \label{fig:sup_correlations_bell}
\end{figure*}

This section discusses the PCC scores attained by MERGE and TRIPLEX across the samples in ST-Net dataset for two cancer-relevant genes - FASN and GNAS. \cref{fig:sup_correlations} shows two bar charts depicting the PCC attained by MERGE and TRIPLEX for each sample in ST-Net. The legend mentions the average PCC over the dataset for each method. It is evident that MERGE achieves a higher average PCC for both the FASN (\(0.39\)) gene and the GNAS (\(0.42\)) gene. Additionally, we can see that there are plenty of samples where MERGE achieves a higher PCC for both genes while TRIPLEX is unable to do so. The reverse however is rarely true. In case of FASN, MERGE performs better than TRIPLEX in \textit{39 samples}, with an average PCC that is higher by \(0.33\). In case of GNAS, in the \textit{41 samples} where MERGE performs better than TRIPLEX, the average PCC achieved by MERGE is higher that that of TRIPLEX by \(0.34\). \cref{fig:sup_correlations_bell} shows the sample-wise bar charts of the PCC achieved by both methods for the two genes. For visual convenience, the vertical red dashed line in each panel splits the chart into two zones. The zone on the left can be considered a \textit{low PCC zone} where a method has achieved a low PCC, less than \(0.25\) for FASN and less than \(0.265\) for GNAS. We can see that the number of samples for this region of the graph is significantly lower for MERGE (blue bars), which means that MERGE exhibits much fewer low PCC samples.

\begin{figure*}[t]
    \centering
    \includegraphics[height=0.9\textheight,keepaspectratio]{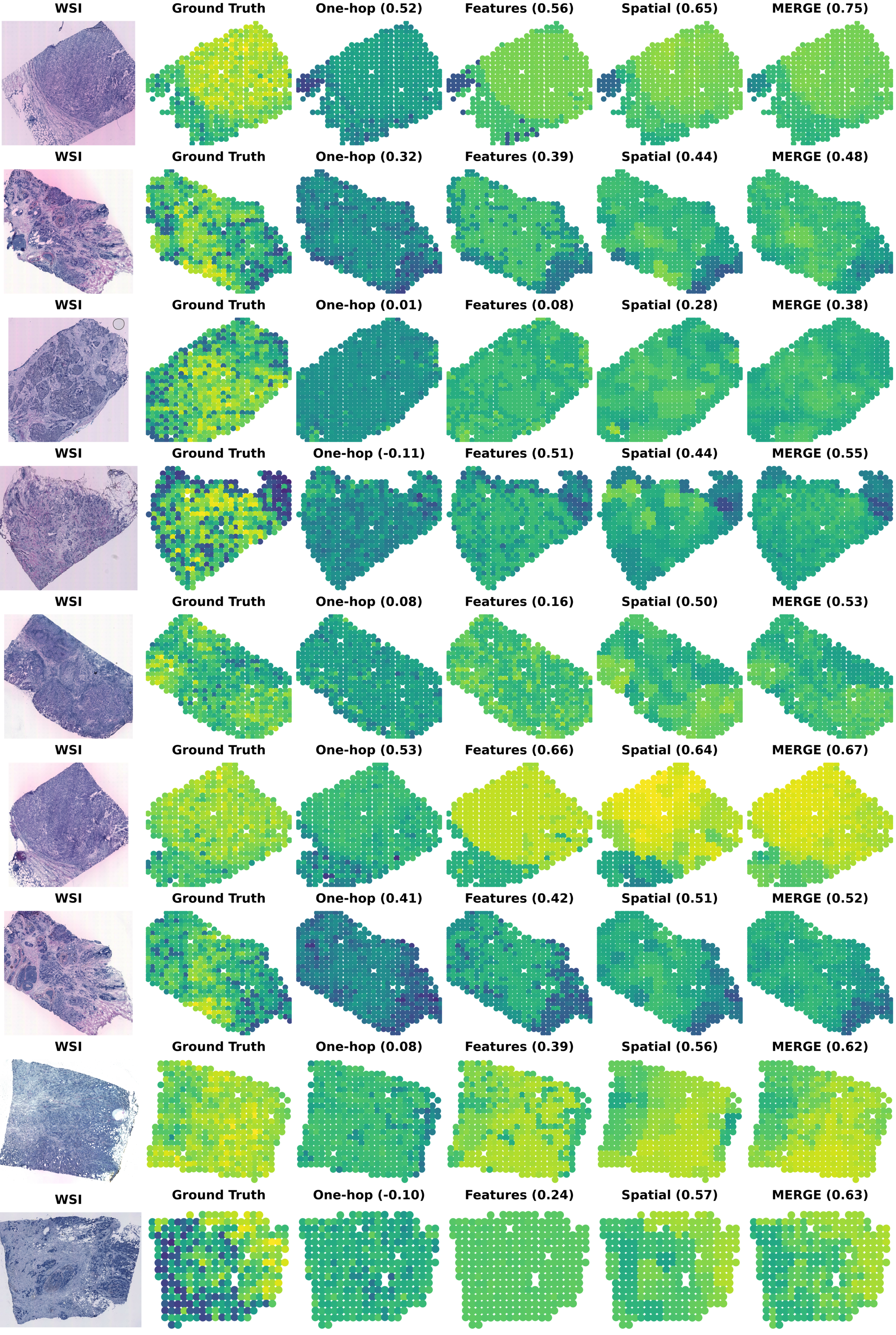}
    \caption{Figure shows PCC between ground truth expressions and predictions for the gene FASN in a few samples. Each row represents a sample, and from the left we have the WSI, the ground truth expressions, predicted expressions using one-hop edges, feature space clustering based edges, spatial clustering based edges and both clustering methods (MERGE).}
    \label{fig:ablation_fasn_0}
\end{figure*}

\begin{figure*}[t]
    \centering
    \includegraphics[height=0.9\textheight,keepaspectratio]{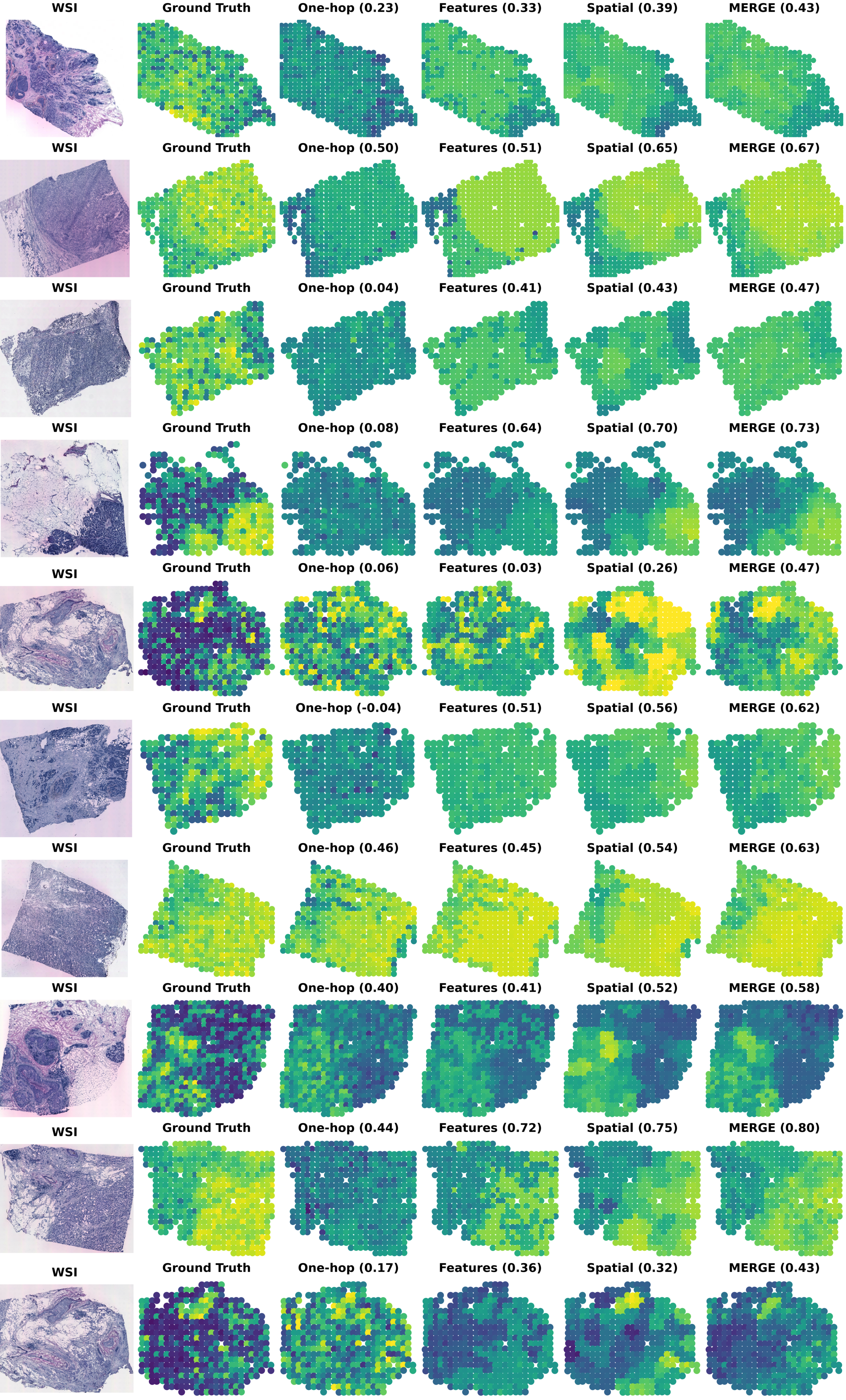}
    \caption{Extension of \cref{fig:ablation_fasn_0}}
    \label{fig:ablation_fasn_1}
\end{figure*}

\begin{figure*}[t]
    \centering
    \includegraphics[height=0.9\textheight,keepaspectratio]{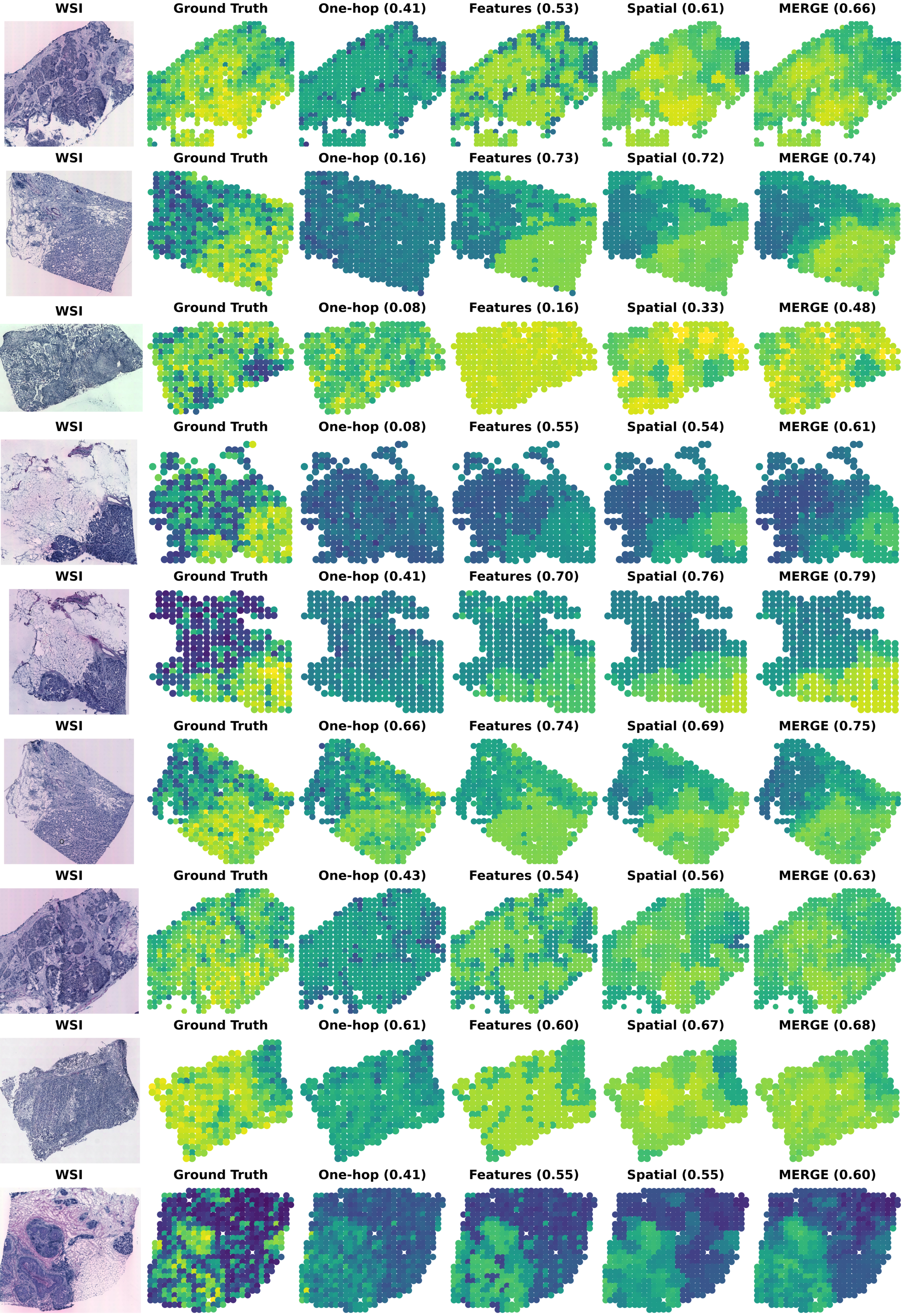}
    \caption{Figure shows PCC between ground truth expressions and predictions for the gene GNAS in a few samples. Each row represents a sample, and from the left we have the WSI, the ground truth expressions, predicted expressions using one-hop edges, feature space clustering based edges, spatial clustering based edges and both clustering methods (MERGE).}
    \label{fig:ablation_gnas_0}
\end{figure*}

\begin{figure*}[t]
    \centering
    \includegraphics[height=0.9\textheight,keepaspectratio]{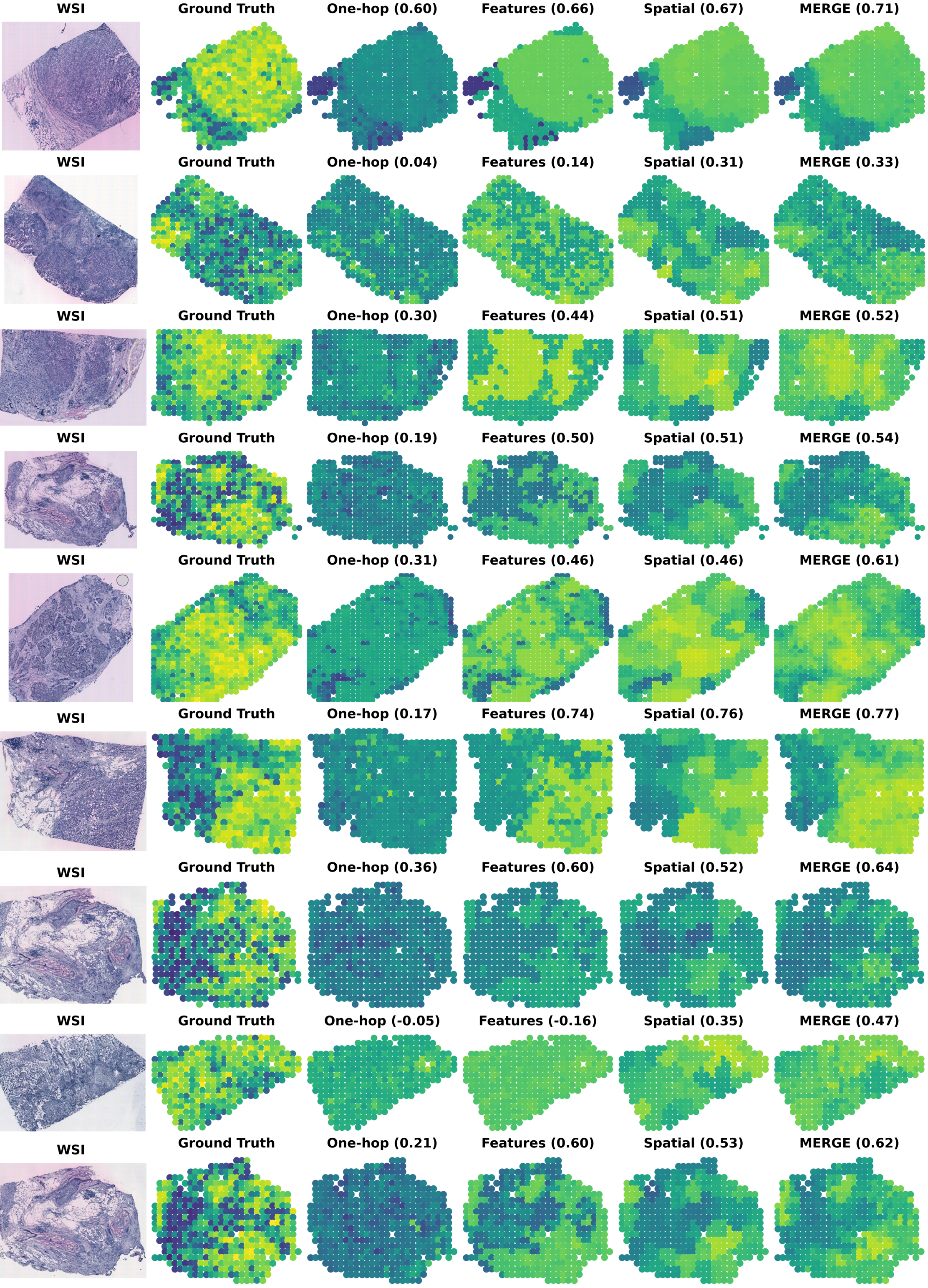}
    \caption{Extension of \cref{fig:ablation_gnas_0}}
    \label{fig:ablation_gnas_1}
\end{figure*}

\subsection{Ablation Visualization}
\label{subsec:sup_ablation_viz}
Using the same ResNet18 based encoder and progressively adding the three modules of our graph construction strategy, we can see a comparative performance of the modules across multiple samples. When we plot the WSI and the ground truth gene expressions alongside the predicted expressions using one-hop edges, feature space clustering-based edges, spatial clustering-based edges, and the combined strategy (MERGE) - we can see that the Pearson Correlation Coefficient of predicted and ground truth gene expressions improves progressively. \cref{fig:ablation_fasn_0} and \cref{fig:ablation_fasn_1} visualize this comparison across multiple samples for the tumor marker FNAS gene. Similarly, \cref{fig:ablation_gnas_0} and \cref{fig:ablation_gnas_1} visualize this comparison across multiple samples for the breast cancer biomarker GNAS gene. For most samples, the PCC achieved using only feature space or spatial clustering is better than that achieved using only one-hop edges. The PCC is highest when using a combination of both clustering methods alongside the one-hop edges.

\clearpage



\end{document}